\documentclass[journal]{IEEEtran}
\usepackage{amssymb} 
\usepackage{newtxtext,newtxmath} 
\usepackage{amsmath,amsfonts}
\usepackage{algorithmic}
\usepackage{algorithm}
\usepackage{array}
\usepackage[caption=false,font=footnotesize,labelfont=sf,textfont=sf]{subfig}
\usepackage{textcomp}
\usepackage{stfloats}
\usepackage{url}
\usepackage{verbatim}
\usepackage{graphicx}
\usepackage{cite}
\usepackage{booktabs} 
\usepackage{hyperref}
\usepackage{multirow}
\usepackage{tabularx}
\usepackage{xcolor}

\begin{document}

\title{TabGRU: An Enhanced Design for Urban Rainfall Intensity  Estimation Using Commercial Microwave Links}

\author{Xingwang Li, Mengyun Chen, Jiamou Liu, Sijie Wang, Shuanggen Jin, Jafet C. M. Andersson, Jonas Olsson, Remco (C. Z.) van de Beek, Hai Victor Habi, Congzheng Han

\thanks{Xingwang Li and Mengyun Chen are with the School of Physics and Electronic Information Engineering, Henan Polytechnic University, Jiaozuo, 454003, China (e-mails: lixingwangbupt@gmail.com; mengyunchen00@gmail.com).

Jiamou Liu and Sijie Wang are with the School of Computer Science, The University of Auckland, Auckland 1010, New Zealand (e-mails: jiamou.liu@auckland.ac.nz; swan387@aucklanduni.ac.nz).

Shuanggen Jin is with the School of Surveying and Land Information Engineering, Henan Polytechnic University, Jiaozuo 454003, China, and also with
Shanghai Astronomical Observatory, Chinese Academy of Sciences, Shanghai 200030, China (e-mail: sgjin@hpu.edu.cn).

Jafet C. M. Andersson, Jonas Olsson, and Remco (C. Z.) van de Beek are with the Swedish Meteorological and Hydrological Institute (SMHI), 601 76 Norrköping, Sweden (e-mails: Jafet.Andersson@smhi.se; Jonas.Olsson@smhi.se; remco.vandebeek@smhi.se).

Hai Victor Habi is with the School of Electrical Engineering, Tel Aviv University, Tel Aviv 69978, Israel (e-mail: haivictorh@mail.tau.ac.il).

Congzheng Han is with the State Key Laboratory of Atmospheric Environment and Extreme Meteorology, Institute of Atmospheric Physics, Chinese Academy of Sciences, Beijing 100029, China (corresponding author: c.han@mail.iap.ac.cn).
}
}

\maketitle
\pagestyle{headings}  
\markboth{Journal of \LaTeX\ Class Files,~Vol.~18, No.~9, September~2020}{}

\begin{abstract}
In the face of accelerating global urbanization and the increasing frequency of extreme weather events, high-resolution urban rainfall monitoring is crucial for building resilient smart cities. Commercial Microwave Links (CMLs) are an emerging data source with great potential for this task. While traditional rainfall retrieval from CMLs relies on physics-based models, these often struggle with real-world complexities like signal noise and nonlinear attenuation. To address these limitations, this paper proposes a novel hybrid deep learning architecture based on the Transformer and a Bidirectional Gated Recurrent Unit (BiGRU), which we name TabGRU. This design synergistically captures both long-term dependencies and local sequential features in the CML signal data. The model is further enhanced by a learnable positional embedding and an attention pooling mechanism to improve its dynamic feature extraction and generalization capabilities. {\color{blue}The model was validated on a public benchmark dataset from Gothenburg, Sweden (June–September 2015). The evaluation used 12 sub-links from two rain gauges (Torp and Barl) over a test period (August 22-31) covering approximately 10 distinct rainfall events. The proposed TabGRU model demonstrated consistent advantages, outperforming deep learning baselines and achieving high coefficients of determination (R$^2$) at both the Torp site (0.91) and the Barl site (0.96). Furthermore, compared to the physics-based approach, TabGRU maintained higher accuracy and was particularly effective in mitigating the significant overestimation problem observed in the PL model during peak rainfall events. This evaluation} confirms that the TabGRU model can effectively overcome the limitations of traditional methods, providing a robust and accurate solution for CML-based urban rainfall monitoring {\color{blue}under the tested conditions}.

\end{abstract}


\section{Introduction}
\IEEEPARstart{A}S global climate change intensifies, extreme precipitation events are occurring more frequently and with greater intensity, {\color{blue}making it crucial to enhance urban resilience. Accurate rainfall estimation is essential for flood warnings and hydrological simulations, directly impacting risk prevention in smart cities}. Traditional rainfall monitoring includes rain gauges, weather radar, and satellite remote sensing. Although rain gauges are accurate, {\color{blue}they suffer from high costs and limited, uneven spatial coverage}\cite{MICHAELIDES2009512}. Weather radar and satellites {\color{blue}offer wide coverage, but their near-surface accuracy is limited}\cite {doviak2006doppler,yates1975meteorological}. With the rapid development of communication technology, {\color{blue}a precipitation monitoring approach} utilizing existing communication network infrastructure is transforming current practices{\color{blue}\cite{messer2006,leijnse2007}}: rainfall intensity retrieval through Commercial Microwave Links (CMLs) networks\cite{han2023potential}. {\color{blue}This technique leverages existing telecommunication infrastructure, offering a low-cost, high-resolution breakthrough in precipitation measurement} \cite{uijlenhoet2018opportunistic,messer2018capitalizing,chwala2019commercial,han2020rainfall}. {\color{blue}Beyond rainfall estimation, recent studies have demonstrated that CMLs can also be utilized to monitor other hydrometeors and atmospheric variables, such as humidity, water vapor fields\cite{david2019analyzing,rubin2023high} and dense fog events\cite{david2013fog,david2018fog}. These studies highlight the potential of CMLs as a versatile, low-cost sensing network for broad hydrometeorological monitoring. The core idea for rainfall retrieval—which remains the most prominent application—is that microwave signals weaken due to scattering and absorption by raindrops. The resulting signal attenuation can be analyzed to estimate path-averaged rainfall intensity}\cite{10064014}, {\color{blue}an idea originally proposed by Atlas in 1977}\cite{atlas1977path}. In 2006, researchers demonstrated that rainfall could be estimated from variations in signal strength between mobile communication base stations\cite{meser2006environmental}, but progress was slowed by the reluctance of communication companies to share data\cite{tollefson2017rain}.
In recent years, {\color{blue}the openness of mobile communication data has spurred practical applications of using microwave signals for rainfall estimation, with experiments initiated in multiple European and African countries. This technique leverages existing telecommunication infrastructure, requiring no dedicated observational equipment, and offers excellent spatial coverage and cost-effectiveness. For example, Dutch scientists constructed a high-precision national rainfall field in 2010 using 2,400 microwave links}\cite{overeem2013country}. {\color{blue}Data from Sweden in 2015 also confirmed that CML measurements are more accurate than conventional radar during flash rainfall events}\cite{tollefson2017rain}. {\color{blue}Due to these advantages, CML technology is particularly suitable for developing countries or mountainous areas that lack dense networks of ground-based observation stations\cite{doumounia2014rainfall,s22093218}. However, this model-driven approach, which is based on physical laws, faces many challenges in practical application despite being interpretable. On the one hand, its effectiveness depends heavily on parameter selection}, and the use of fixed parameters makes it difficult to generalize across different link lengths and complex geographic environments\cite{regonesi2019limitations,han2021characteristics}. On the other hand, the model usually assumes that rainfall is uniformly distributed over the link paths, which ignores the spatial randomness and inhomogeneity of rainfall in real situations, leading to large estimation errors and reduced accuracy.

Against this backdrop, recent studies have increasingly adopted machine learning (ML) and deep learning (DL) techniques to enhance the automatic extraction and modeling of complex rainfall time-series features. {\color{blue}While this data-driven paradigm bypasses explicit physical modeling and has shown great potential, a closer review of existing approaches reveals several persistent and recurring challenges. First, models often face a trade-off between scalability and complexity. As early as 2013, Wu and Chau (2013) first constructed a rain prediction model using Support Vector Regression (SVR) and a local neural network}\cite{wu2013prediction}. {\color{blue}However, the training time for SVR increases exponentially with the sample size, making it unsuitable for large-scale rainfall datasets. In contrast, Ye and Gao (2022) proposed a method based on a multi-scale Convolutional Neural Network (CNN)}\cite{ye2022msstnet}, {\color{blue}which captures features at different spatial scales using multi-scale convolutional kernels. Although it can effectively extract spatiotemporal information, the use of 3D convolutions also leads to a significant increase in computational load. Second, many architectures struggle to capture long-term temporal dependencies. Shi and Chen (2015) proposed the Convolutional Long Short-Term Memory (ConvLSTM) model}\cite{shi2015convolutional}, {\color{blue}which improves the ability of the traditional Long Short-Term Memory (LSTM) to handle spatiotemporal data by introducing a convolutional structure. However, its inherent Recurrent Neural Network (RNN) architecture causes prediction errors to accumulate with increasing time steps, limiting its accuracy in predicting long-term sequence trends. Other architectures, such as simple two-layer LSTMs}\cite{pudashine2020deep} {\color{blue}or BiLSTMs}\cite{yu2023attention}, {\color{blue}remain structurally shallow or possess limited memory capacity, failing to capture deeper temporal features in complex rainfall sequences. Third, a critical limitation lies in robustness and generalization. Habi and Messer (2019) conducted an in-depth study on RNN-based models for rain detection (i.e., wet-dry classification)}\cite{9020603}. {\color{blue}Experimenting with datasets from two different climatic regions, Israel and Sweden, they found that the Gated Recurrent Unit (GRU) is more efficient than traditional LSTM units and also revealed a key shortcoming of deep learning models in this domain: the models struggle to generalize to previously unseen climatic regions. Building on this, Habi and Messer (2020) further proposed an RNN model for directly estimating rain rate and compared it with the traditional power-law (PL) model across three aspects: accuracy, robustness, and complexity}\cite{habi2020recurrent}. {\color{blue}The research clearly indicates that although the RNN model surpasses the traditional method in accuracy, its robustness is poorer. Finally, some novel approaches lack sufficient validation. For example, Sergey Timinsky (2023) explored a rain prediction model based on Cycle-Consistent Generative Adversarial Networks (CycleGAN)}\cite{timinsky2023rain}, {\color{blue}training it with unpaired datasets of attenuation measurements and rain gauge observations. However, their experiments used only the Root Mean Square Selection Error (RMSSE) for performance evaluation and lacked a comparative analysis with other rain estimation methods, making it difficult to fully validate the model's effectiveness and reliability. In summary, while data-driven methods show great promise as a new paradigm, existing models often force a compromise: they are either too simple (failing to capture complex dynamics), too complex (requiring high computational cost), or lack proven robustness and generalization for this specific domain.}

To address these limitations, we propose a deep learning model, named TabGRU, to accurately estimate the regional rain rate and improve prediction accuracy. For traditional RNNs, when dealing with long sequences, it is difficult to capture the dependence of long-distance time steps due to the vanishing gradient problem\cite{info14110598}. In contrast, the Transformer can directly compute the correlation between any two time steps through self-attention\cite{luo2023self}, which completely solves this problem. This is important for time-dependent rainfall sequence prediction, as a rainfall event may span many time steps. The TabGRU model demonstrates superior performance in both efficiency and accuracy across rainfall estimation tasks. It effectively captures both long-term trends and short-term fluctuations in rainfall data by leveraging its unique learnable positional encoding, Multi-Head Attention, and attention pooling. The main findings and contributions of this paper are as follows:

(1) In this study, the advantages of CMLs in high-resolution rainfall estimation are verified in an urban area where microwave links are densely distributed, and a hybrid architecture for rainfall estimation based on Transformer-BiGRU is proposed, which utilizes the existing microwave link data to achieve the estimation of regional rainfall intensity.

(2) A BiGRU layer is introduced to make up for the inadequacy of Transformer in capturing local time-series patterns, and the advantages of the two complement each other, so that the model in this paper can focus on global features while also learning the local time-series patterns of the sequences, which improves the model's prediction ability for short-term rainfall events.

(3) We introduce customizable learnable positional encoding to dynamically adjust the representation of the positional encoding of the input sequences to enhance the flexibility and generalization ability of the model. An attention pooling layer is introduced to enable the model to adaptively assign weights to each position of the input sequence to improve the model's accuracy.

(4) We conduct a comprehensive experimental validation, which shows that TabGRU significantly outperforms not only a range of deep learning baselines but also a traditional model-driven approach, demonstrating its superior performance and robustness across diverse rainfall scenarios.

This paper is organized as follows: Section II presents the technical background of rainfall estimation. Section III presents an overview of the study area and dataset. Section IV describes the overall architecture of the model in this paper. Section V analyzes the experimental results, and Section VI discusses the findings and concludes the experimental results.

\section{Rainfall estimation methods}
\subsection{Traditional rainfall estimation methods}
Traditional rainfall estimation techniques are usually categorized into two types: ground monitoring and space measurement\cite{sun2018review}.
Ground-based precipitation observation instruments include tipping bucket rain gauges, weighing rain gauges, laser raindrop spectrometers and others, which are capable of providing highly accurate point measurements. Space remote sensing measurements, such as weather radar and meteorological satellites, make up for the lack of spatial coverage of ground-based observations and are able to enable precipitation monitoring over large regions or even the whole country. However, different measurement methods are based on different principles and have their own advantages and disadvantages in terms of their accuracy, cost, and applicability to different scenarios. To facilitate comparison, the mainstream traditional rainfall estimation techniques are summarized in Table \ref{tab:methods}.

\begin{table*}[htbp]
\centering
\caption{Comparison of Different Rainfall Measurement Methods}
\label{tab:methods}
\begin{tabular}{p{1.2cm}|p{1cm}|p{5cm}|p{4cm}|p{4.5cm}}
\hline 
\textbf{Method} & \textbf{Type} & \textbf{Working Principle} & \textbf{Advantages} & \textbf{Disadvantages/Limitations} \\
\hline\hline 
\multirow{4}{*}{\parbox{1.2cm}{\raggedright Direct Ground Measurements}} & 
Tipping Bucket Rain Gauge & When accumulated rainwater reaches a preset volume, the bucket tips, triggering a counter that records tips to estimate the amount of rainfall\cite{Gou2025ComparativeAnalysis}. & It features a simple structure, low cost, and high reliability\cite{segovia2023tipping}. & It suffers from a tipping-threshold effect, which leads to inaccurate monitoring of trace rainfall and the exact start and end times of rain events\cite{QHXZ202405012}. \\
\cline{2-5} 
& Weighing Rain Gauge & Measures real-time changes in the weight of water in a container using a weighing sensor to calculate rainfall amount\cite{COLLI2014186}. & No triggering threshold; It can effectively capture trace precipitation and the precise start and end times of rain events. & Complex structure and high cost; Sensitive to environmental factors, with higher maintenance requirements\cite{Liang2024Comparative}. \\
\cline{2-5}
& 
Laser Disdrometer & Based on laser occultation, it infers the size and fall velocity of individual raindrops from changes in the laser-beam signal to determine rainfall intensity\cite{Pu2010Characteristics}. & It has high temporal resolution and can be used for fine-grained analysis of precipitation structure. & Expensive; not suitable for large-scale deployment, and susceptible to interference from wind, aerosols, and other factors\cite{Du2017Observational}. \\
\hline 
\multirow{6}{*}{\parbox{1.2cm}{\raggedright Remote Sensing Indirect Measurements}} 
& Weather Radar & Using the Z–R relationship, it estimates rainfall rate by transmitting electromagnetic pulses and analyzing the returned reflectivity from precipitation particles\cite{li2016application}. & Enables regional monitoring, compensating for the limitations of point measurements. & Its signal is susceptible to interference; It may have larger errors when estimating near-surface rainfall\cite{raich2018vertical}; It relies on ground-based data for calibration. \\
\cline{2-5}
& Satellite Remote Sensing & Uses thermal-infrared or microwave sensors to indirectly estimate rainfall intensity: thermal-infrared methods rely on cloud-top temperature, while microwave methods rely on particle emission/scattering mechanisms\cite{li2016application}. & Capable of global precipitation monitoring with extensive spatial coverage\cite{yates1975meteorological}. & Spatiotemporal resolution is often low; Satellite retrievals are subject to instrumental and physical limitations, and they tend to have larger estimation errors for precipitation beneath cloud tops\cite{Statusofsatellite}. \\
\hline 
\end{tabular}
\end{table*}

\subsection{{\color{blue}The model-driven approach for CMLs}}
Although traditional rainfall monitoring techniques have been widely used for precipitation monitoring, they each have deficiencies in terms of spatial and temporal resolution, deployment cost, or estimation accuracy. With the continuous development of communication network infrastructure, commercial microwave links are gradually becoming a new type of distributed meteorological sensor with great potential in the construction of smart cities\cite{han2020rainfall}. The technology has significant advantages, such as low cost, high density, and real-time monitoring, which provide a new development direction for surface rainfall monitoring. In studies on rainfall estimation based on CMLs, the physics-based approaches are mostly based on the Power-Law (PL) model. Its theoretical roots lie in the physical law of electromagnetic wave interaction with precipitation particles in the atmosphere. Specifically, when the microwave signal propagates in the near-surface atmosphere, the raindrops on the path will have absorption and scattering effects on it, resulting in signal energy attenuation. A power-law relationship exists between rain-induced attenuation and rainfall intensity, and we can calculate the rain-induced attenuation using the model provided by the International Telecommunication Union (ITU-R)\cite{regonesi2019limitations}. It is calculated as shown in Equation (\ref{eq:power_law}) :
\begin{equation}
A = kL = {\color{blue}aR^b L}
\label{eq:power_law} 
\end{equation}
where $A$ is the attenuation induced by rainfall (in dB); k is the specific rain attenuation (in dB/m). $L$ is the length of the microwave link (in m); $R$ is the rainfall intensity (in mm/h); {\color{blue}The $a$ and $b$ are empirical parameters that depend on the signal's frequency and polarization, which are related to the Raindrop Size Distribution (DSD)}\cite{1141845}. 

\subsection{{\color{blue}Limitations of the model-driven approach}}
{\color{blue}However, the practical application of the model-driven approach is complicated by several factors that introduce significant uncertainty. First,} DSD is the distribution of the number of raindrops of different sizes per unit volume, which is a natural factor affecting the uncertainty of the relationship between rain attenuation and rainfall intensity. Different types of precipitation, such as convective heavy precipitation and stratiform light precipitation, have very different raindrop spectral characteristics, which also correspond to different a and b values. Under complex weather conditions, when it is difficult to accurately model DSD variations or when local DSD observations are lacking, power-law models usually choose generic parameter values based on frequency and polarization mode. For microwave links over short paths or in environments with weak precipitation and strong local convection, using generic parameter values can lead to significant errors\cite{YIN2015Research}. Second, the physical inversion method requires processes such as dry and wet time period discrimination and removal of the Wet Antenna Attenuation (WAA). In this regard, the WAA refers to rainwater adhering to the surface of the radome to form a water film or water droplets, which can cause additional attenuation unrelated to rainfall\cite{fencl2020atmospheric}. This is the most challenging aspect of the model-driven approach and one of the main sources of error. Since the WAA is associated with a variety of factors such as rainfall intensity, wind speed, radome material, and has highly nonlinear characteristics, which together result in a complex attenuation mechanism\cite{8762206}, it is difficult to establish an accurate physical model for removal. The main advantage of the model-driven approach is that the model is highly interpretable, as it corresponds to a specific physical process in each processing link, and does not require a large amount of historical observation data for training. However, the model's heavy reliance on physical processes induces systematic deviations that are unquantifiable under complex rainfall scenarios due to its preset parameters. Throughout the physical inversion process, errors may exist at each step, from wet/dry discrimination to baseline estimation and finally to the WAA. These errors can be transmitted and accumulated step by step, thus affecting the overall accuracy of the rainfall intensity inversion results. The physical baseline model implemented in this paper follows this classic processing workflow.

\subsection{The data-driven approach as a solution}
To overcome the limitations of the model-driven approach mentioned above, in recent years, researchers have started to introduce data-driven approaches to automatically learn the mapping between the Received Signal Level (RSL) of the microwave link and the rainfall intensity from large-scale link signal datasets and real-time precipitation data using deep learning models\cite{shen2018transdisciplinary}. In particular, these deep learning models, such as the CNN, RNN, and LSTM, are widely used to model the dynamic association between link signal sequences and rainfall intensity due to their superior ability to process time-series data. The core idea is to construct an end-to-end model that directly learns the complex nonlinear mapping between the time series of the RSL of a microwave link and the measured rainfall intensity. Rather than the multi-step processing flow in traditional physical models, this approach treats the entire inversion process as a black box\cite{reichstein2019deep}. The model is trained on large volumes of paired CML signal and rainfall intensity data, and automatically learns the nonlinear relationship between the two. Unlike model-driven methods, which struggle to accurately remove WAA, deep learning models can operate without relying on preset physical functions. By learning the unique characteristics of changes in CML signals at the beginning and end of rainfall, these models extract temporal and statistical patterns associated with WAA, thereby achieving implicit modeling of it.
. This approach allows the model to better adapt to various complex rainfall scenarios, endowing it with greater predictive accuracy and robustness. The introduction and development of deep learning methods are key to promoting the transition of CML-based rainfall estimation from experimental research to operational applications. Such progress is indispensable for constructing more accurate and robust hydrometeorological monitoring systems for smart cities in the future.

\section{Data and Experiment}
\subsection{Study area}
\begin{figure}[htbp]
    \centering
    \includegraphics[width=3.5in]{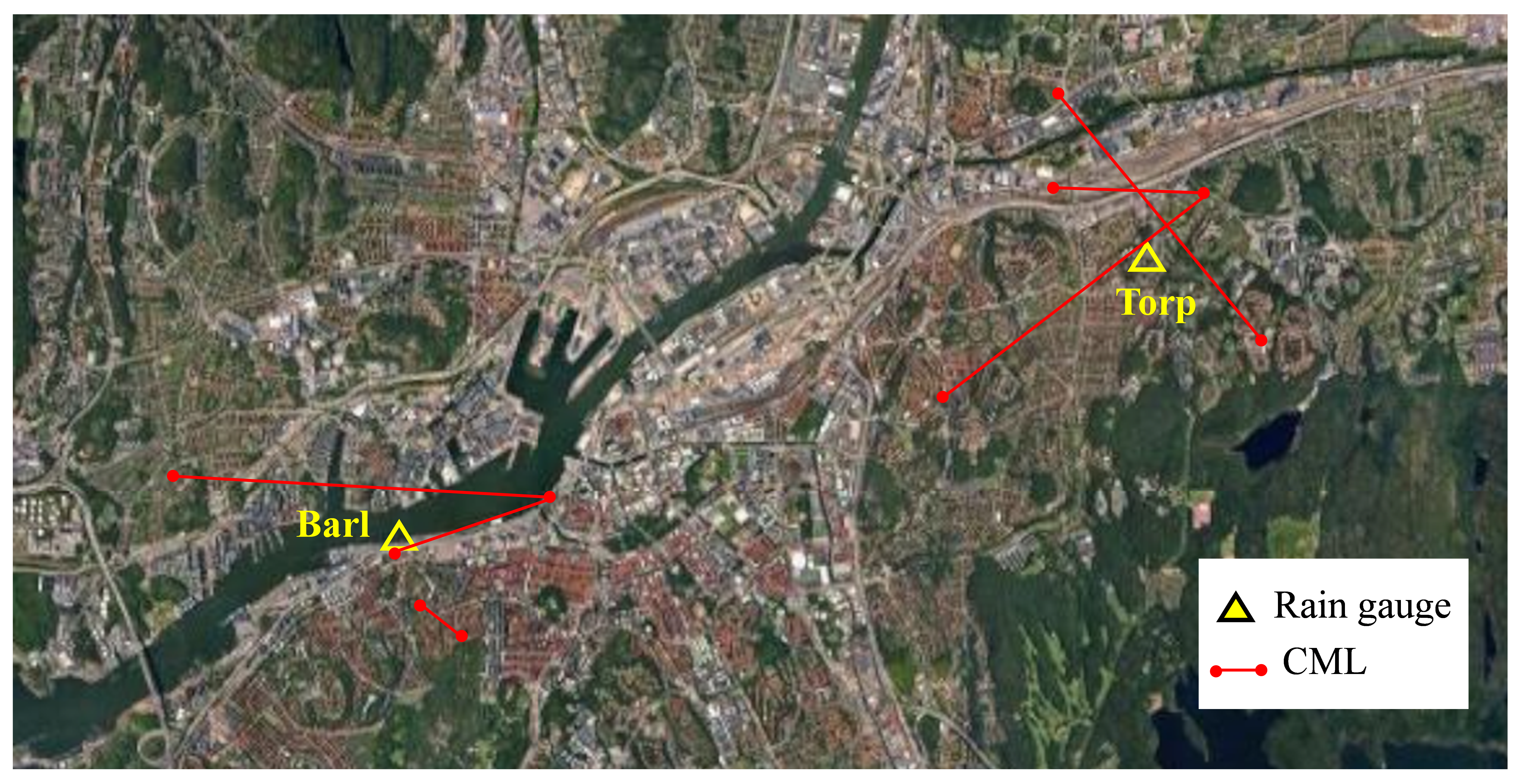}

    \caption{{\color{blue}Distribution of links and rain-gauge stations.}}
    \label{fig:map}
\end{figure}

The study area of this research encompasses the city of Gothenburg, Sweden, and its surrounding regions. The city is located at 57\textdegree 42$'$N latitude and 11\textdegree 58$'$E longitude on the southwestern coast of Sweden. Gothenburg has a temperate oceanic climate with abundant precipitation throughout the year, resulting in an annual rainfall of between 700 and 800 mm. Because of the relatively uniform distribution of precipitation, there is no obvious distinction between dry and rainy seasons. This feature provides abundant monitoring data for microwave link rainfall prediction studies. {\color{blue}The distribution of the links and rain gauges utilized in this study is presented in Figure \ref{fig:map}}.

\subsection{Dataset}
\begin{figure}[htbp]
    \centering
    \includegraphics[width=3.5in]{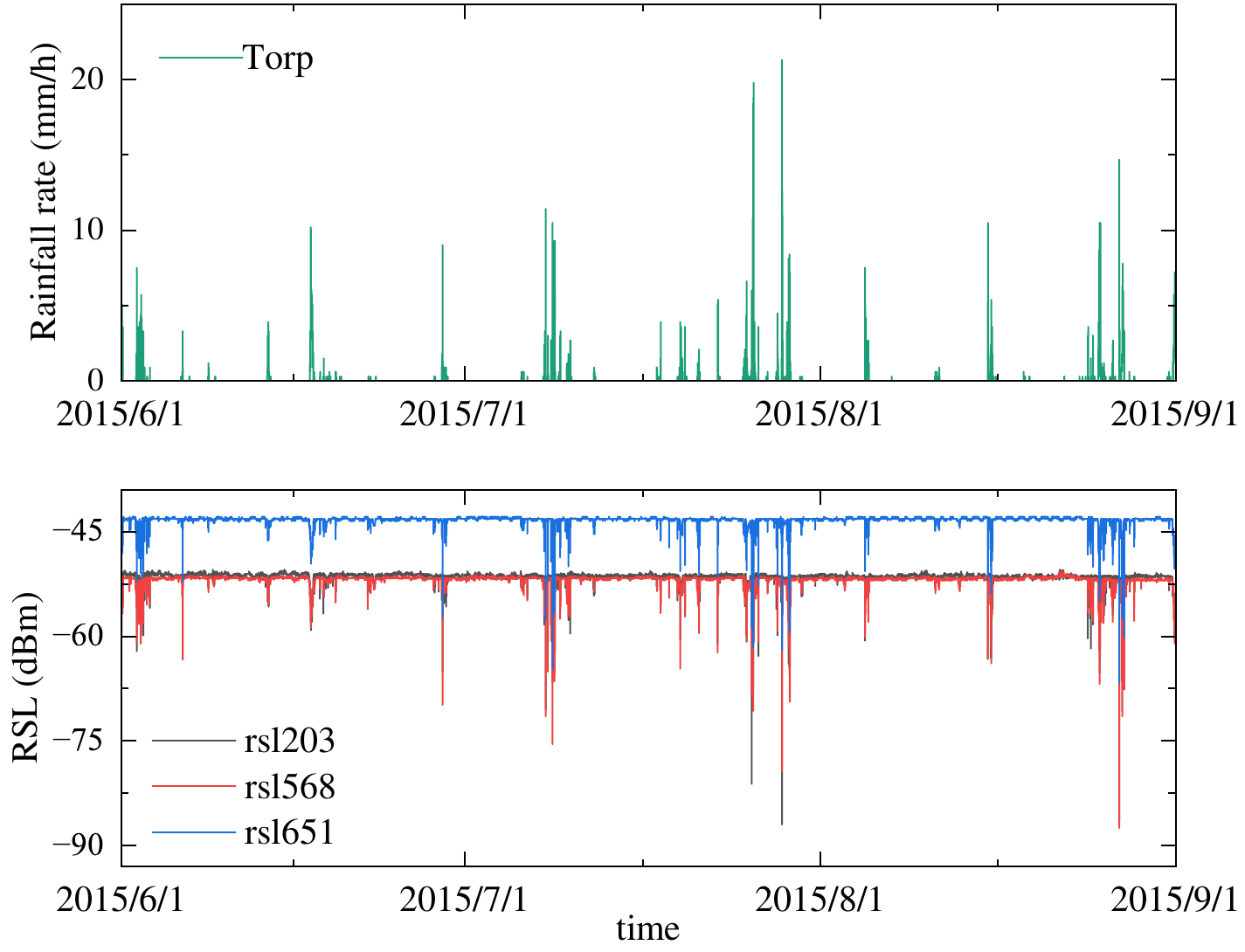}

    \caption{Time series of rainfall rate measured by the Torp rain gauge (top) and 1-minute averaged RSL from three microwave links (bottom) during June–August 2015.}
    \label{fig:1}
\end{figure}

The dataset comprises authentic measurements from both CMLs and urban rain gauges in Gothenburg and its vicinity during the period from June 1 to August 31, 2015\cite{andersson2022openmrg}. The link data {\color{blue}include the Transmitted Signal Level (TSL) at the transmitter end}, the RSL at the receiver end, {\color{blue}the operating frequency, the link length}, and the coordinates of the near (transmitter) and far (receiver) nodes for each sub-link {\color{blue}(defined as the signal in one direction of a bidirectional link)}. {\color{blue}For this study, the RSL was downsampled from 10-s to 1-min resolution by averaging six consecutive 10-s samples, ensuring temporal alignment with the rain-gauge observations. The rain gauges used are the Torp and Barl tipping-bucket gauges, each with a measurement resolution of 0.1 mm and a sampling interval of 1 minute.} Figure \ref{fig:1} {\color{blue}illustrates the relationship between the 1-minute RSL and rainfall intensity for three representative links (IDs 203, 568, and 651) during June–August 2015. To quantify their relationship, the Pearson correlation coefficient (PCC) between the RSL and the corresponding rain-gauge rainfall rate was computed after temporal alignment. The results show a clear negative correlation, with PCC values of $-0.65$ (Link 203), $-0.68$ (Link 568), and $-0.65$ (Link 651), indicating that RSL decreases systematically with increasing rainfall intensity. This confirms that signal attenuation measured by CMLs is strongly correlated with precipitation intensity.}

\subsection{Data preprocessing}
The data preprocessing pipeline for this study comprised two primary stages: downsampling input features derived from CMLs and processing the rain gauge target variable, including unit conversion and smoothing. First, to address mismatched temporal resolutions among heterogeneous data sources, we temporally aligned the input features. The raw CML data, at 10-second resolution, were downsampled to match the rain gauge's 1-minute resolution. This was achieved by averaging non-overlapping windows of six consecutive 10-second RSL samples to produce a 1-minute RSL series synchronized to the rain-gauge timestamps. Second, the model's targets were processed. Because the rain gauge provides 1-minute accumulations, we converted these to hourly rainfall rates (mm/h) by multiplying by 60 and then smoothed the result using a moving-average filter to reduce noise. The smoothed hourly rainfall rate was used as the model target. To ensure a rigorous evaluation while preserving the temporal integrity of the data for this time-series task, we employed a chronological split with buffer periods. Specifically, the training set comprises data from June 1 to August 2, the validation set from August 4 to August 20, and the test set from August 22 to August 31. The one-day buffers between these partitions serve to prevent data leakage by ensuring single rainfall events are not split across sets. To preserve local trends, missing values (NaNs) were imputed using polynomial interpolation so that the imputed values follow local changes in the original series\cite{lepot2017interpolation}. Neural network models, such as GRU, Transformer, and LSTM, are trained via gradient-based optimization methods to learn optimal parameters. During optimization, differences in feature scale can cause gradients to have vastly different magnitudes across parameter directions, which may slow or destabilize training. To mitigate this, input features were scaled using the RobustScaler method\cite{technologies9030052}, as defined in Equation(\ref{eq:scaling}):
\begin{equation}
x_{\text{scaled}} = \frac{x - \operatorname{median}(x)}{\mathit{IQR}(x)}
\label{eq:scaling}
\end{equation}
where $IQR=Q_3-Q_1$ is the interquartile range, $Q_3$ is the third quartile and $Q_1$ is the first quartile; $median(x)$ is the median.

\subsection{Experimental Setup}
The performance of the model proposed in this paper is highly dependent on the choice of key hyperparameters. {\color{blue}A critical hyperparameter is the input time-series length. In all experiments, this length was uniformly set to 30 minutes. This choice was made primarily to ensure the model can capture the key physical characteristics of a minimal effective rainfall event, as 30 minutes is a common threshold used to define independent rainfall events \cite{atmos14091322}.} The key hyperparameters of the model can be divided into three categories: model architecture parameters (such as the number of multi-head attention heads, the number of network layers), regularization parameters (such as Dropout rate), and training configuration parameters (such as batch size, learning rate). The finalized optimal hyperparameter combination is detailed in Table \ref{tab:hyperparameters}.
\begin{table}[htbp]
\centering
\caption{Key hyperparameters of the proposed model.}
\label{tab:hyperparameters}
\renewcommand{\arraystretch}{1} 
\begin{tabularx}{0.7\columnwidth}{l | >{\centering\arraybackslash}X} 
\hline
\textbf{Hyperparameter}      & \textbf{Value} \\ \hline\hline
Transformer Encoder Layers   & 3              \\ \hline
Number of Attention Heads    & 4              \\ \hline
Learning Rate                & 5e-4           \\ \hline
GRU Hidden Units             & 64             \\ \hline
GRU Layers                   & 1              \\ \hline
Dropout Rate                 & 0.3            \\ \hline
Window Size                  & 30             \\ \hline
Epochs                       & 150            \\ \hline
\end{tabularx} 
\end{table}

\subsection{Evaluation Metrics}
For the rainfall prediction results, four widely used metrics are employed to evaluate the performance of the proposed model: Root-Mean-Square error (RMSE), Coefficient of Determination ($R^2$), Pearson Correlation Coefficient (PCC), and Mean Absolute Error (MAE). The formulas are as follows:
\begin{align}
RMSE &= \sqrt{\frac{1}{n}\sum_{i=1}^{n}(y_i - \hat{y}_i)^2} \label{eq:rmse} \\
R^2 &= 1 - \frac{\sum_{i=1}^{n}(y_i - \hat{y}_i)^2}{\sum_{i=1}^{n}(y_i - \bar{y})^2} \label{eq:r2} \\
PCC &= \frac{\sum_{i=1}^{n}(y_i - \bar{y})(\hat{y}_i - \bar{\hat{y}})}{\sqrt{\sum_{i=1}^{n}(y_i - \bar{y})^2} \cdot \sqrt{\sum_{i=1}^{n}(\hat{y}_i - \bar{\hat{y}})^2}} \label{eq:pcc} \\
MAE &= \frac{1}{n}\sum_{i=1}^{n}|y_i - \hat{y}_i| \label{eq:mae}
\end{align}
Where $y_i$ is the i-th true value, $\hat{y}_i$ is the i-th predicted value, $n$ is the number of samples, $\bar{y}$ is the mean of the true values, and $\overline{\hat{y}}$ is the mean of the predicted values. Smaller RMSE and MAE indicate better predictive performance, whereas larger $R^2$ and PCC values indicate better performance.

\section{Model design}
\subsection{Feature engineering}
This study focuses on urban rainfall monitoring, for which {\color{blue}the Torp and Barl rain gauges} were selected as the ground-truth references. {\color{blue}These gauges were chosen as they} provide an ideal combination of high data reliability and strategic locations. According to the OpenMRG dataset\cite{andersson2022openmrg}, {\color{blue}both are weighing-type gauges, which are} more reliable than the network's tipping-bucket gauges {\color{blue}(prone to biases). Furthermore, both gauges are situated in areas} with high CML network density, making {\color{blue}them} representative and optimal testbeds for validating our method. {\color{blue}To ensure a comprehensive analysis and avoid selection bias, we adopted a rigorous, objective selection criterion. We selected three bidirectional links located within a 1 km radius of each of our reference rain gauges, yielding a total of 6 bidirectional links. Furthermore, and crucially, we utilized the RSL sequences from both unidirectional sub-links for all 6 links. This resulted in a total of 12 sub-links being included in the experiment. The specific physical parameters of these selected links are detailed in Table \ref{tab:cml_params}}.
\begin{table}[htbp]
\centering
\caption{Detailed physical parameters of the CMLs selected for this study.}
\label{tab:cml_params}
\begin{tabular}{l|c|c|c}
\hline
\textbf{Link ID} & \textbf{Length (km)} & \textbf{Frequency (GHz)} & \textbf{\begin{tabular}[c]{@{}c@{}}Sampling Interval\\ (s)\end{tabular}} \\ \hline\hline
{\color{blue}201}      & {\color{blue}2.79}             & {\color{blue}32.26}                   & {\color{blue}10}                                                                      \\ \hline
203                   & 2.79                          & 32.45                                & 10                                                                      \\ \hline 
{\color{blue}568}      & {\color{blue}2.52}             & {\color{blue}32.42}                   & {\color{blue}10}                                                                      \\ \hline
{\color{blue}567}      & {\color{blue}2.52}             & {\color{blue}33.23}                   & {\color{blue}10}                                                                      \\ \hline
651                   & 1.28                          & 38.32                                & 10                                                                      \\ \hline
{\color{blue}652}      & {\color{blue}1.28}             & {\color{blue}37.32}                   & {\color{blue}10}                                                                      \\ \hline
{\color{blue}435}      & {\color{blue}0.25}             & {\color{blue}37.27}                   & {\color{blue}10}                                                                      \\ \hline
{\color{blue}436}      & {\color{blue}0.25}            & {\color{blue}38.53}                   & {\color{blue}10}                                                                      \\ \hline
{\color{blue}587}      & {\color{blue}1.26}             & {\color{blue}32.45}                   & {\color{blue}10}                                                                      \\ \hline
{\color{blue}588}      & {\color{blue}1.26}             & {\color{blue}33.26}                   & {\color{blue}10}                                                                      \\ \hline
{\color{blue}601}      & {\color{blue}3.05}             & {\color{blue}33.15}                   & {\color{blue}10}                                                                      \\ \hline
{\color{blue}602}      & {\color{blue}3.05}             & {\color{blue}32.33}                   & {\color{blue}10}                                                                      \\ \hline
\end{tabular}
\end{table}

To determine the model's input features, we first analyzed the signal characteristics of these three links. We confirmed that the TSL remained essentially constant throughout the entire study period. This characteristic is consistent with the official description of the OpenMRG dataset, which states that due to disabled Adaptive Transmit Power Control (ATPC), the TSL signal is constant for a majority (68\%) of the sub-links\cite{andersson2022openmrg}. When the TSL is constant, variations in path attenuation caused by rainfall are directly reflected by changes in the RSL. Consequently, the dynamics of RSL can directly represent such signal attenuation. Based on this analysis, and to directly leverage the physical relationship between the raw signal and rainfall while simplifying the model input, we ultimately selected the RSL time series from the three links as the primary input features for the deep learning model. The model's learning target was provided by the 1-minute rainfall intensity data measured by the Torp rain gauge.

 In this paper, the temporal features extracted from the raw data timestamps are encoded by sine and cosine functions as shown in Equation (\ref{eq:sin}) and Equation (\ref{eq:cos}). This periodic encoding avoids representing time as discrete integers, as by mapping timestamps to sine and cosine components\cite{bansal2025temporal}, it better captures diurnal periodicity and is more consistent with the underlying physical processes. This treatment {\color{blue}is crucial for helping the model learn the diurnal (daily) patterns of signal attenuation. It enables the model to better distinguish true precipitation-induced attenuation (e.g., from afternoon thunderstorms, which typically have a distinct onset time) from non-precipitation attenuation (such as nighttime dew), as both phenomena can exhibit strong daily cycles.}
\begin{align}
x_{\text{sin}} &= \sin\left(\frac{2\pi \cdot t}{T}\right) \label{eq:sin} \\
x_{\text{cos}} &= \cos\left(\frac{2\pi \cdot t}{T}\right) \label{eq:cos}
\end{align}
where $t$ is the  temporal unit (hours or minutes); $T$ is the period of this unit, e.g., $24$ (hours) and $60$ (minutes).

\subsection{Overall framework}
\begin{figure*}[!htbp] 
  \centering
  \includegraphics[width=0.8\linewidth,keepaspectratio]{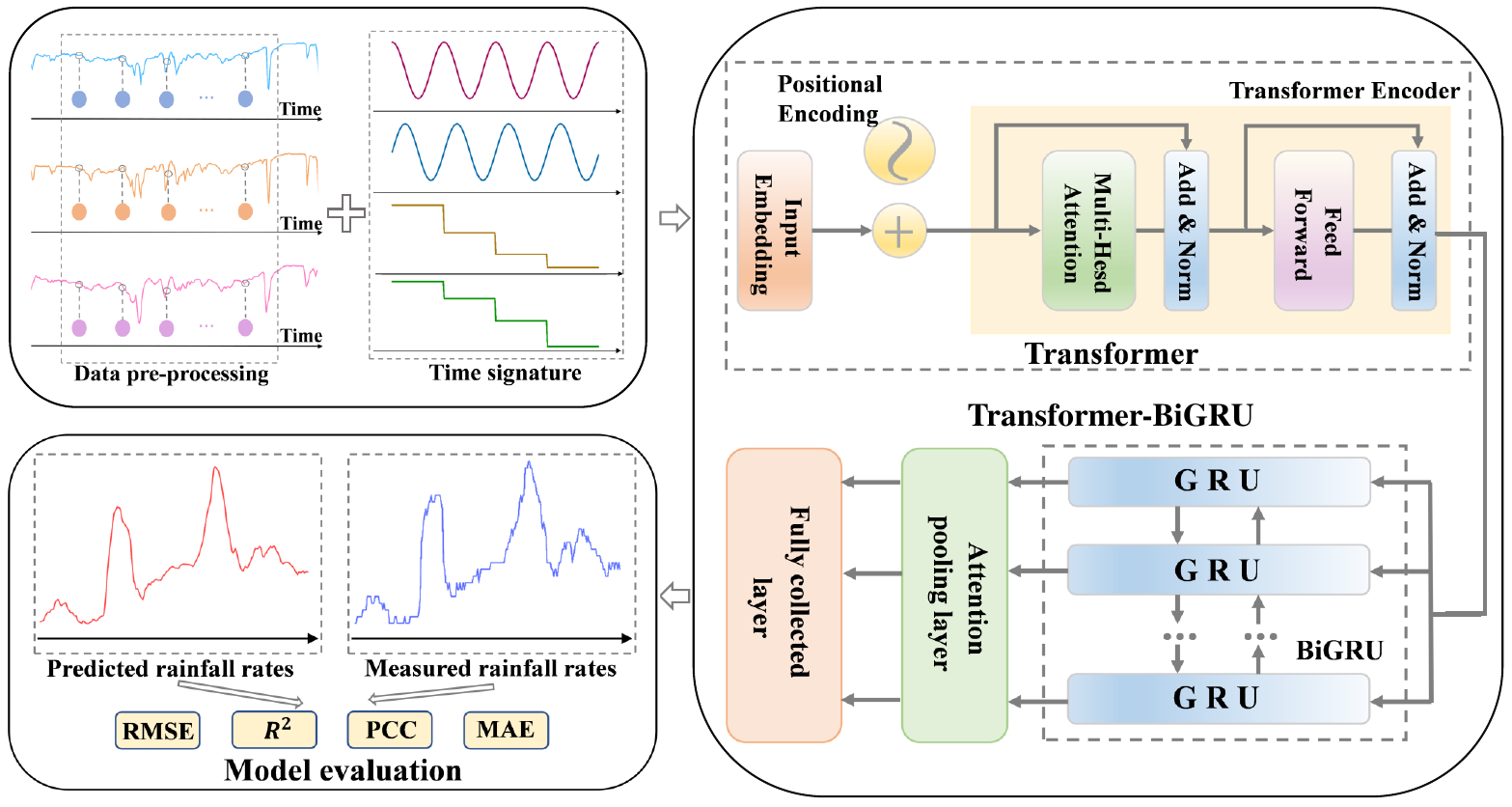}
  \caption{Model Structure Diagram}
  \label{fig:modelarch}
\end{figure*}

To cope with the challenges of parameter dependence and the modeling dilemma of the wet antenna effect in the rainfall inversion method based on CMLs for a model-driven approach, this paper proposes a hybrid deep neural network model, which aims to explore the potential of CML-based data to improve the accuracy and practicality of rainfall estimation. The model adopts a hierarchical learning strategy, and the overall framework is shown in Figure.\ref{fig:modelarch}. The model integrates Transformer, BiGRU, and attention pooling to achieve comprehensive modeling of the temporal features of rainfall.

The workflow of the model is as follows: firstly, the RSL time series from the three links and their associated temporal features are linearly projected into a higher-dimensional feature space to enhance their feature representation; subsequently, a Learnable Positional Encoding (LPE) mechanism is introduced to dynamically inject temporal information into the input sequences. The positionally encoded features are first processed by a multi-layer Transformer encoder to capture multi-scale temporal patterns and long-term rainfall trends. The Transformer's outputs are then fed into a BiGRU, which, by virtue of its bidirectional gated-recurrent structure, models local and short-term dynamics to capture minute-level rainfall fluctuations. Finally, an attention pooling layer aggregates the BiGRU outputs via learned weights, and a fully connected layer maps the aggregated features nonlinearly to predict the next minute rainfall intensity. This model combines the Transformer's global modeling capability with the BiGRU's ability to capture local time-series dynamics, thereby satisfying the dual requirement for multi-scale temporal dependence in rainfall estimation.

\subsection{Multi-scale timing modeling based on Transformer and BiGRU}
\subsubsection{Transformer}
The Transformer model is a deep learning model architecture for Natural Language Processing (NLP), whose core strength is the self-attention mechanism\cite{vaswani2017attention}. In the model proposed in this paper, the Transformer encoder component is used as a module for capturing long-term dependencies and extracting global features. For rainfall data with complex periodic and seasonal features, the multi-head self-attention mechanism can compute the intrinsic correlation between any two time steps in the sequence in parallel, irrespective of the temporal distance between them. This not only significantly reduces the training time but also enables the model to capture difficult long-term dependencies and nonlinear relationships more efficiently, which is especially suitable for analyzing long series of rainfall data over months or even years.

Since the Transformer's self-attention mechanism cannot perceive the temporal order of the rainfall sequence, this paper equips the model with a learnable position encoding module. This module generates learnable position encodings to inject position information into each time step of the input sequence — ensuring the model recognizes the temporal structure in the data. This module enables the model to dynamically adjust the position encoding according to the characteristics of the actual data during training via back-propagation. Unlike traditional fixed coding based on sine and cosine functions\cite{liu2020learning}, learnable positional encodings are implemented as trainable parameters optimized during training. In this module, we define a position coding matrix by the following Equation(\ref{eq:learnable_matrix}):
\begin{equation}
    P_{\mathrm{learnable}} \in \mathbb{R}^{1 \times L \times d_{\mathrm{model}}}
    \label{eq:learnable_matrix}
\end{equation}
where $L$ is the length of the input sequence and $d_{\text{model}}$ is the feature dimension of each time step. This position encoding matrix is a learnable parameter that can be optimized through the model's backpropagation process, dynamically adjusting according to task needs. The input feature sequence is represented as $X = [x_1, x_2, \dots, x_L]$ ( where each $x_i \in \mathbb{R}^d$ ). The model then combines the optimized position encoding with this sequence step-by-step to obtain the final input representation, as shown in Equation(\ref{eq:feature_update}).
\begin{equation}
    X' = X + P_{\mathrm{learnable}}
    \label{eq:feature_update}
\end{equation}
Among them, $P_{\mathrm{learnable}}$ is the learnable positional encoding that has been optimized during training, which ensures that the model adapts its positional encoding to the temporal characteristics of the actual data. With the learnable positional-encoding layer, a position vector is added to the input at each time step; this helps the Transformer—which does not inherently encode positional order—both recognize temporal structure and retain its parallelizability.

The multi-head self-attention mechanism lies at the core of the model for capturing long-term dependencies in rainfall data. Rainfall processes are often influenced by long-lived, large-scale weather systems, producing long-term dependencies that can persist for several hours. Traditional RNNs struggle to capture such long-range dependencies over these time scales. By contrast, self-attention can directly compute correlation weights between any two time steps (positions) in the sequence, giving the model a global view that enables it to detect long-term patterns. Moreover, the multi-head design allows different attention heads to focus on different aspects of the input and mitigates biases from any single head\cite{cui2024superiority}. The computational process is expressed in Equation (\ref{eq:multihead})(\ref{eq:head}).
\begin{gather}
    \mathrm{MultiHead}(Q,K,V) = \mathrm{Concat}(\mathrm{head}_1, \ldots, \mathrm{head}_h)W^O \label{eq:multihead} \\
    \mathrm{head}_i = \mathrm{Attention}(QW_i^Q, KW_i^K, VW_i^V) \label{eq:head}
\end{gather}
where $\mathrm{head}_i$ is the output of the ith attention head; $h$ is the number of heads; $W^O \in \mathbb{R}^{h d_v \times d_{\mathrm{model}}}$ is the output mapping matrix; $W_i^Q \in \mathbb{R}^{d_{\mathrm{model}} \times d_k}$, $W_i^K \in \mathbb{R}^{d_{\mathrm{model}} \times d_k}$, $W_i^V \in \mathbb{R}^{d_{\mathrm{model}} \times d_v}$, $d_k = d_v = d_{\mathrm{model}}/h$. For the rainfall estimation task, multi-head self-attention not only extracts multi-perspective features and learns correlations among different features, but also attends to all historical time steps within the window at each prediction time, thereby preserving key historical information in the rainfall process.

\subsubsection{BiGRU}
\begin{figure}[!htbp] 
  \centering
  \includegraphics[width=0.9\linewidth,height=0.4\textheight,keepaspectratio]{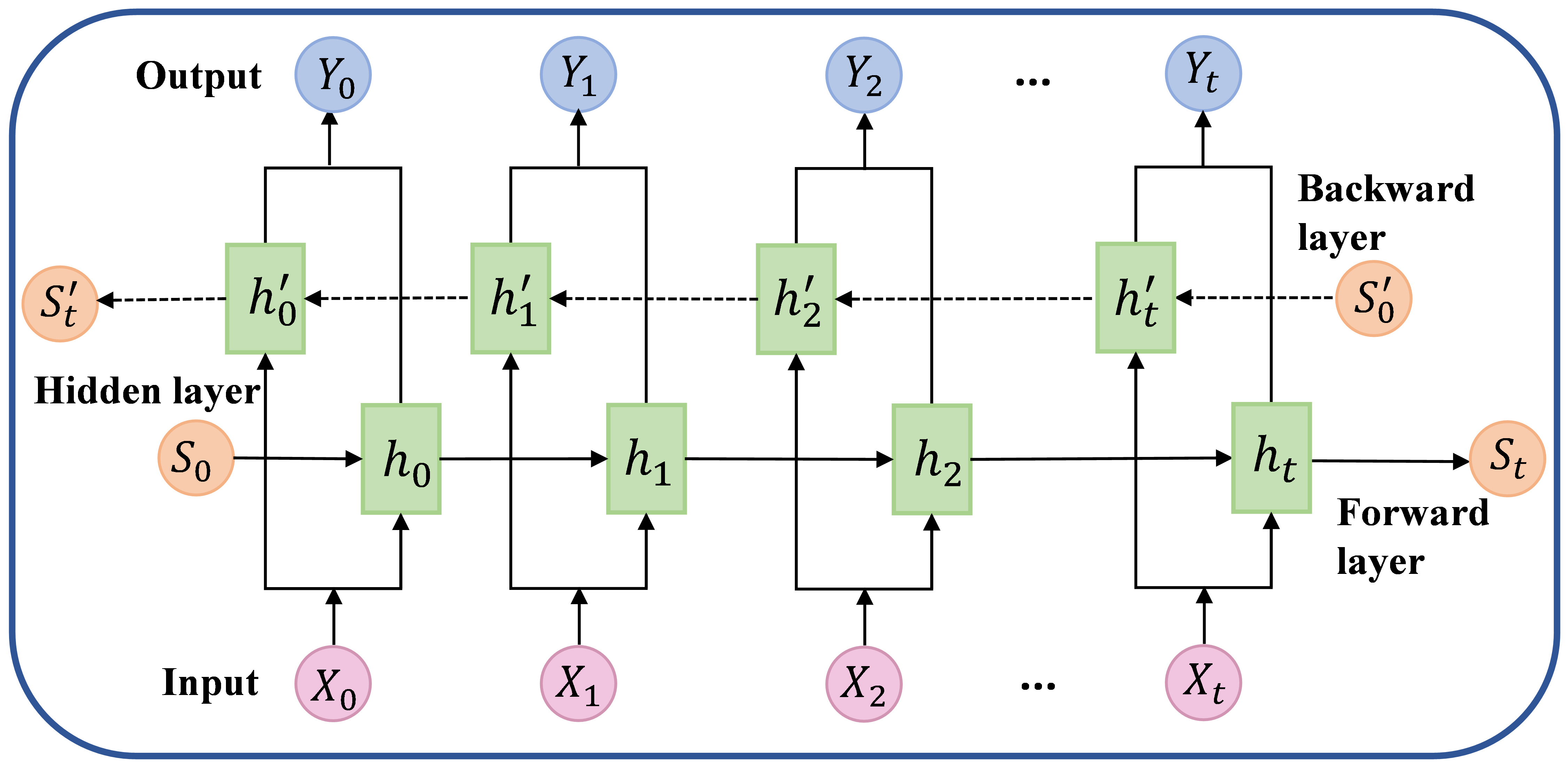} 
  \caption{Structural Diagram of Bidirectional Gated Recurrent Unit (BiGRU)} 
  \label{fig:bigruarch}
\end{figure}

The GRU is a variant of RNN specialized for processing sequence data \cite{chung2014empirical}. In this paper, we use a BiGRU to process the feature sequences output by the Transformer encoder, capturing short-term rainfall fluctuations. A BiGRU consists of two GRU layers that process the sequence in opposite directions. This bidirectional modeling incorporates context from both past and future time steps. Its structure is shown in Figure.\ref{fig:bigruarch}.
Where $h_t$ represents the forward GRU layer,$h_{t}^{\prime}$ represents the backward GRU layer, $S_t$ and $S_{t}^{\prime}$ represent the hidden states, $X_t$ is the input layer, and $Y_t$ is the output layer. The BiGRU module complements the Transformer's global attention by modeling local temporal dynamics through gated recurrent units and a bidirectional structure, helping to mitigate bias and capture minute-level fluctuations.

\subsubsection{Attention Pooling}
Not all moments in a rainfall event are equally important. For example, periods of dramatic signal change near the rainfall peak clearly contain more valuable information for future rainfall estimation than periods of steady or light rainfall{\color{blue}\cite{zhang2022rap}}. To avoid losing or diluting this critical information, we introduce an attention pooling module. This module automatically learns and assigns attention weights based on the hidden state at each time step \cite{zafar2022comparison}, enabling the model to adaptively focus on key time segments in the rainfall event and to assign them higher weights during aggregation. The formulation is shown in Equation(\ref{eq:exp}):
\begin{equation}
\alpha_t = \frac{exp(f(x_t))}{\sum_{i = 1}^{L} exp(f(x_i))}
\label{eq:exp}
\end{equation}
Here, $\alpha_t$ is the attentional weight of the t-th position; $f(x_t)$ is the score function applied to the features at position t. After obtaining the attention weights for all positions, the pooled result is computed by weighted summation as shown in Equation(\ref{eq:weightedsum}). 
\begin{equation}
    Z = \sum\limits_{t=1}^{L} {\color{blue}\alpha_t}  x_{t}
    \label{eq:weightedsum}
\end{equation}
where $Z$ is the pooled vector, which is a weighted average of the features at each time step of the input sequence. Compared with taking the output of the last time step directly or using other pooling methods, attention pooling does not discard critical time segments of the sequence. This approach allows the model to automatically learn and emphasize the time-step information most useful for rainfall estimation while preserving the temporal integrity of the input, thereby improving the accuracy and robustness of the model.

\section{Results and discussions}
{\color{blue}To rigorously validate the generalization capability of our proposed method, we independently conducted two complete sets of experiments based on the two reference rain gauges, Torp and Barl. First, it is necessary to quantify the effective rain events within the datasets. Defining rainfall events using metrics such as the Minimum Inter-event Time (MIT) is not only physically meaningful but also widely applied in the analysis of rainfall intermittency \cite{dunkerley2015intra}. Building upon this established event-based framework, we follow the established methodology in Han (2021)\cite{han2021}, an effective rain event is defined as having a rainfall rate $>$ 0.1 mm/h, a duration of at least 30 minutes, and a rain-free period of at least 1 hour separating it from the next event. Based on this definition, the number of effective rain events identified in the rainfall series for each of the two gauges is detailed in Table \ref{tab:Rain}.}
\begin{table}[htbp]
  \centering
  \caption{{\color{blue}Statistics of effective rainfall events identified from the Torp and Barl rain gauge time series}}
  \begin{tabular}{c|c|c|c|c}
    \hline
    \parbox[c]{1.5cm}{\centering {\color{blue}Rain Gauge}} & \parbox[c]{1cm}{\centering {\color{blue}Training Events}} & \parbox[c]{1cm}{\centering {\color{blue}Validation Events}} & \parbox[c]{1cm}{\centering {\color{blue}Test Events}} & \parbox[c]{1cm}{\centering {\color{blue}Total Events}} \\
    \hline\hline
    {\color{blue}Torp} & {\color{blue}36} & {\color{blue}6} & {\color{blue}9} & {\color{blue}51} \\
\hline
    {\color{blue}Barl} & {\color{blue}38} & {\color{blue}5} & {\color{blue}11} & {\color{blue}54} \\
\hline
  \end{tabular}
  \label{tab:Rain}
\end{table}
\subsection{Overall performance and stability analysis}
{\color{blue}To comprehensively assess the effectiveness and generalization ability of the proposed TabGRU model, this study performs rigorous comparative analyses across two independent experimental sites (Torp and Barl), each equipped with three bidirectional microwave links (six links in total). All models are trained and evaluated under the same data splitting protocol to ensure the fairness of comparisons.}

{\color{blue}First, at the Torp site, a statistical analysis of the ground-truth (Gauge) data was performed to provide an objective baseline for the model's error metrics, as summarized in Table \ref{tab:Torpall}. The test set rainfall exhibited a mean of 0.28 mm/h and a high standard deviation of 1.12 mm/h, revealing the highly variable nature of the rainfall process, which poses a substantial challenge for accurate prediction. On this dataset, the TabGRU model achieved an RMSE of 0.34 mm/h, considerably lower than the data's own standard deviation, and an MAE of 0.07 mm/h. The R$^2$ and PCC values reached 0.91 and 0.96, respectively. Compared to the strongest baseline model (BiGRU), TabGRU reduced the RMSE and MAE by 17.07\% and 11.11\%, respectively, while also improving the R$^2$ and PCC by 4.59\% and 3.23\%.}
\begin{table}[htbp]
\centering
\caption{\textcolor{blue}{Statistical Analysis of Rainfall Intensity Predicted by Various Models and That Measured by the Torp Rain Gauge}}
\begin{tabular}{c|c|c|c|c}
\hline
\parbox[c]{1.5cm}{\centering \textcolor{blue}{Model}} &
\parbox[c]{1cm}{\centering \textcolor{blue}{RMSE}} &
\parbox[c]{1cm}{\centering \textcolor{blue}{$R^2$}} &
\parbox[c]{1cm}{\centering \textcolor{blue}{PCC}} &
\parbox[c]{1cm}{\centering \textcolor{blue}{MAE}} \\
\hline\hline

\textcolor{blue}{RNN} & \textcolor{blue}{0.56} & \textcolor{blue}{0.77} & \textcolor{blue}{0.89} & \textcolor{blue}{0.10} \\
\hline

\textcolor{blue}{Transformer} & \textcolor{blue}{0.48} & \textcolor{blue}{0.82} & \textcolor{blue}{0.91} & \textcolor{blue}{0.10} \\
\hline

\textcolor{blue}{GRU} & \textcolor{blue}{0.41} & \textcolor{blue}{0.86} & \textcolor{blue}{0.93} & \textcolor{blue}{0.09} \\
\hline

\textcolor{blue}{BiGRU} & \textcolor{blue}{0.41} & \textcolor{blue}{0.87} & \textcolor{blue}{0.93} & \textcolor{blue}{0.09} \\
\hline

\textcolor{blue}{Trans-GRU} & \textcolor{blue}{0.43} & \textcolor{blue}{0.85} & \textcolor{blue}{0.93} & \textcolor{blue}{0.09} \\
\hline

\textcolor{blue}{TabGRU} & \textcolor{blue}{\textbf{0.34}} & \textcolor{blue}{\textbf{0.91}} & \textcolor{blue}{\textbf{0.96}} & \textcolor{blue}{\textbf{0.08}} \\
\hline
\end{tabular}
\label{tab:Torpall}
\end{table}

\textcolor{blue}{To further validate the model's generalization ability, the same experiment was replicated at the Barl site (see Table \ref{tab:Barlall}). An objective baseline for the error metrics at the Barl site was established through a similar statistical analysis of its test set ground-truth data. The results showed a rainfall mean of 0.29 mm/h and a standard deviation of 1.16 mm/h, likewise indicating high variability in the rainfall process. As shown in Table \ref{tab:Barlall}, the TabGRU model's RMSE was 0.25 mm/h, lower than the data's standard deviation, with an MAE of 0.06 mm/h. The model's R$^2$ and PCC reached 0.96 and 0.98, respectively. Compared to the strongest baseline model at this site (Trans-GRU), TabGRU reduced the RMSE by 37.5\% and the MAE by 33.33\%, while improving the R$^2$ by 9.09\% and the PCC by 4.26\%. This data indicates that TabGRU consistently reduced prediction errors and improved goodness-of-fit across both experimental sites.}
\begin{table}[htbp]
  \centering
  \caption{\textcolor{blue}{Statistical Analysis of Rainfall Intensity Predicted by Various Models and That Measured by the Barl Rain Gauge}}
  \begin{tabular}{c|c|c|c|c}
    \hline
    \parbox[c]{1.5cm}{\centering \textcolor{blue}{Model}} &
    \parbox[c]{1cm}{\centering \textcolor{blue}{RMSE}} &
    \parbox[c]{1cm}{\centering \textcolor{blue}{$R^2$}} &
    \parbox[c]{1cm}{\centering \textcolor{blue}{PCC}} &
    \parbox[c]{1cm}{\centering \textcolor{blue}{MAE}} \\
    \hline\hline

    \textcolor{blue}{RNN} & \textcolor{blue}{0.71} & \textcolor{blue}{0.64} & \textcolor{blue}{0.82} & \textcolor{blue}{0.14} \\
    \hline

    \textcolor{blue}{Transformer} & \textcolor{blue}{0.53} & \textcolor{blue}{0.79} & \textcolor{blue}{0.91} & \textcolor{blue}{0.13} \\
    \hline

    \textcolor{blue}{GRU} & \textcolor{blue}{0.47} & \textcolor{blue}{0.84} & \textcolor{blue}{0.91} & \textcolor{blue}{0.10} \\
    \hline

    \textcolor{blue}{BiGRU} & \textcolor{blue}{0.44} & \textcolor{blue}{0.85} & \textcolor{blue}{0.93} & \textcolor{blue}{0.11} \\
    \hline

    \textcolor{blue}{Trans-GRU} & \textcolor{blue}{0.40} & \textcolor{blue}{0.88} & \textcolor{blue}{0.94} & \textcolor{blue}{0.09} \\
    \hline

    \textcolor{blue}{TabGRU} &
    \textcolor{blue}{\textbf{0.25}} &
    \textcolor{blue}{\textbf{0.96}} &
    \textcolor{blue}{\textbf{0.98}} &
    \textcolor{blue}{\textbf{0.06}} \\
    \hline
  \end{tabular}
  \label{tab:Barlall}
\end{table}

\begin{figure}[htbp]
  \centering
  \subfloat[\textcolor{blue}{RMSE}]{
    \includegraphics[width=0.48\columnwidth]{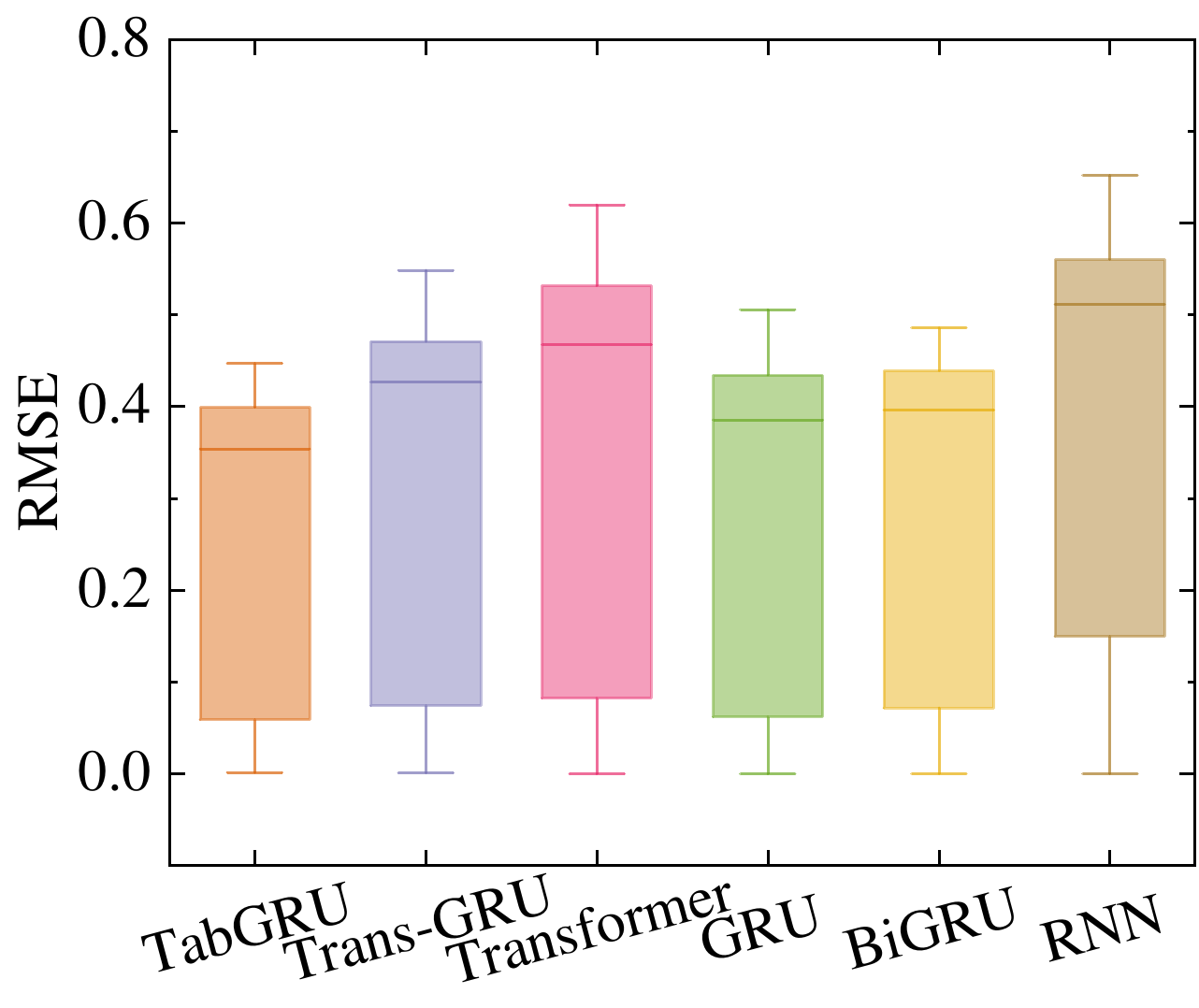}
    \label{fig:Torp_RMSE}}
  \hfill
  \subfloat[\textcolor{blue}{$R^2$}]{
    \includegraphics[width=0.48\columnwidth]{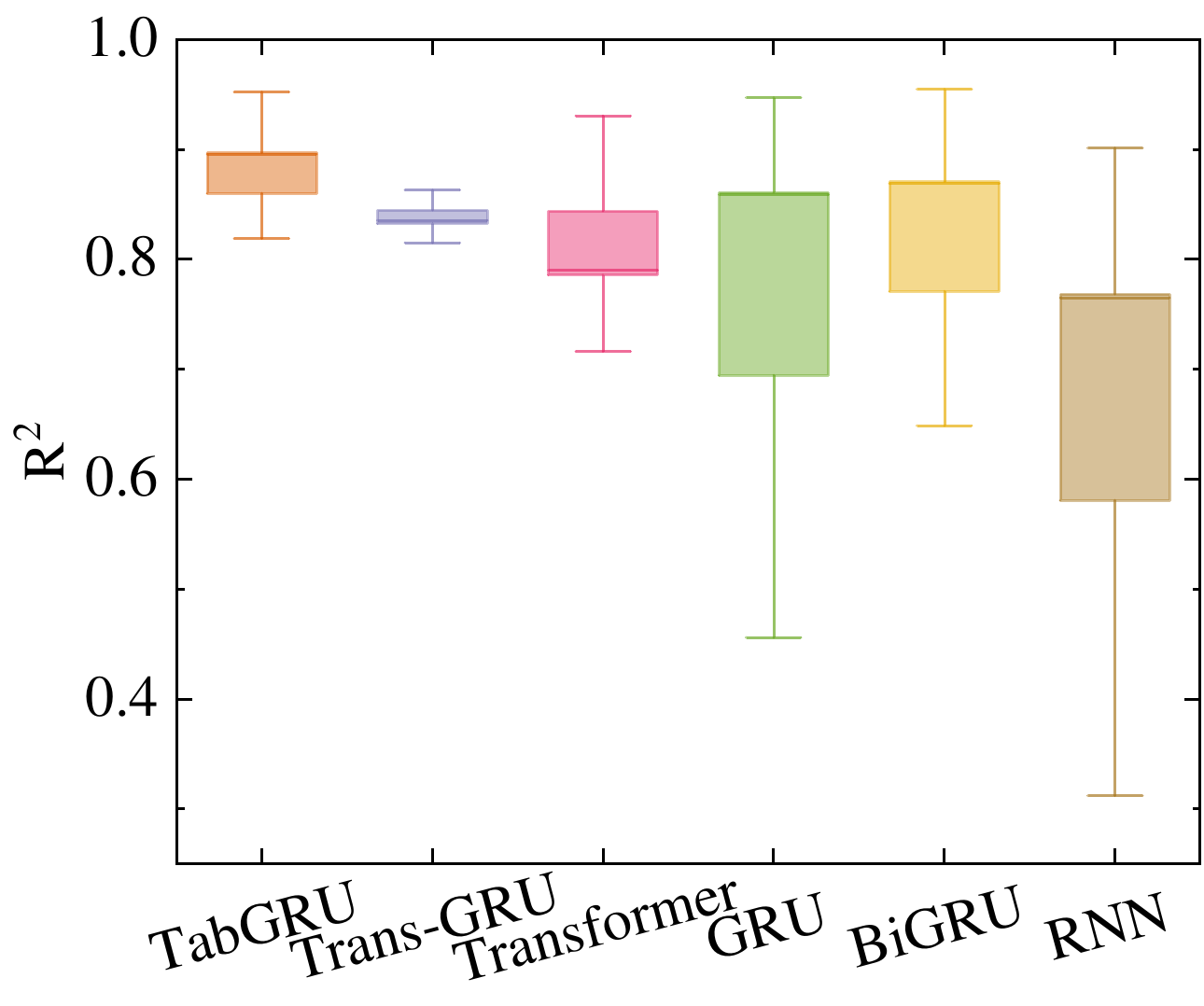}
    \label{fig:Torp_R2}}
  
  \vspace{0.2cm}
  
  \subfloat[\textcolor{blue}{PCC}]{
    \includegraphics[width=0.48\columnwidth]{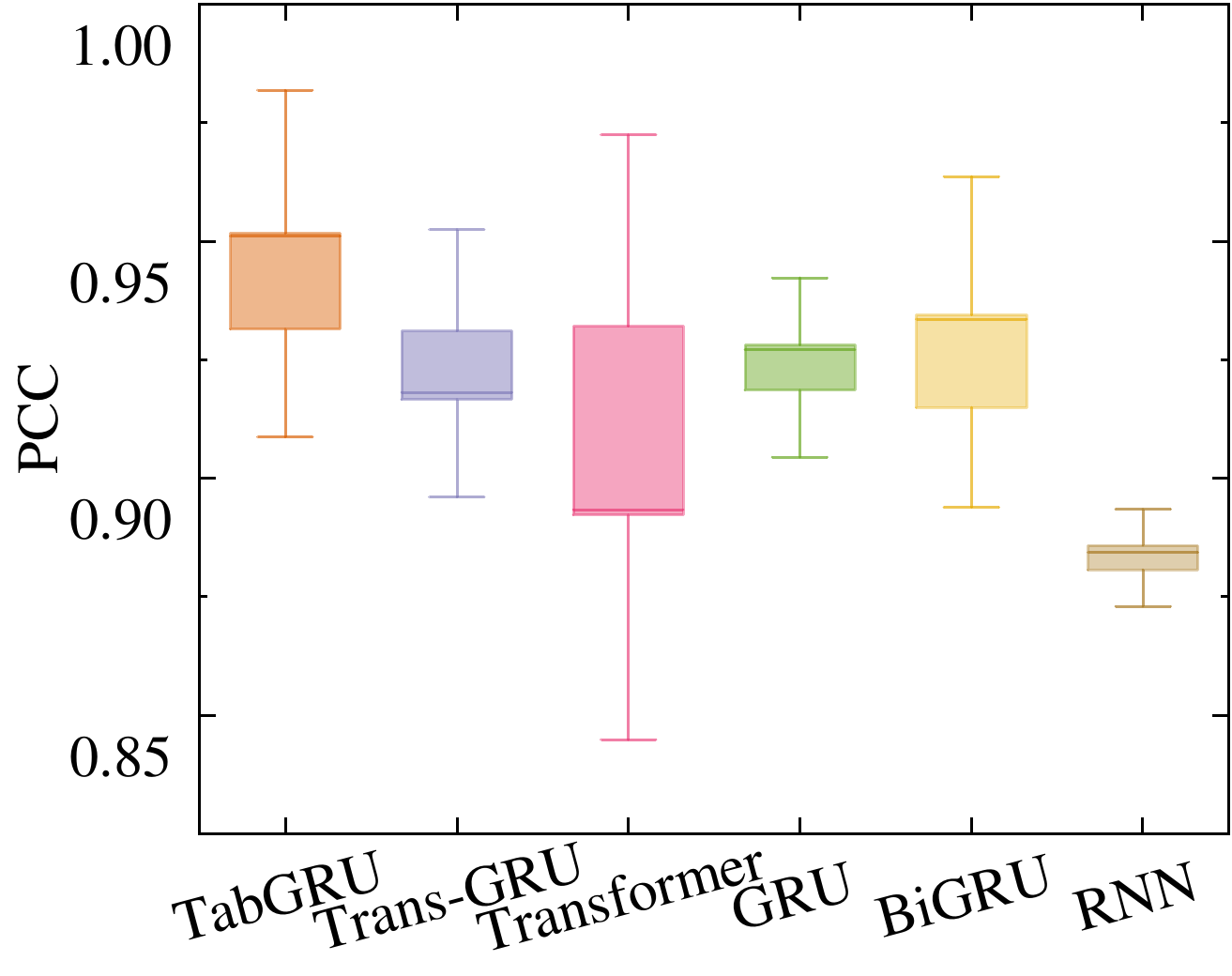}
    \label{fig:Torp_PCC}}
  \hfill
  \subfloat[\textcolor{blue}{MAE}]{
    \includegraphics[width=0.48\columnwidth]{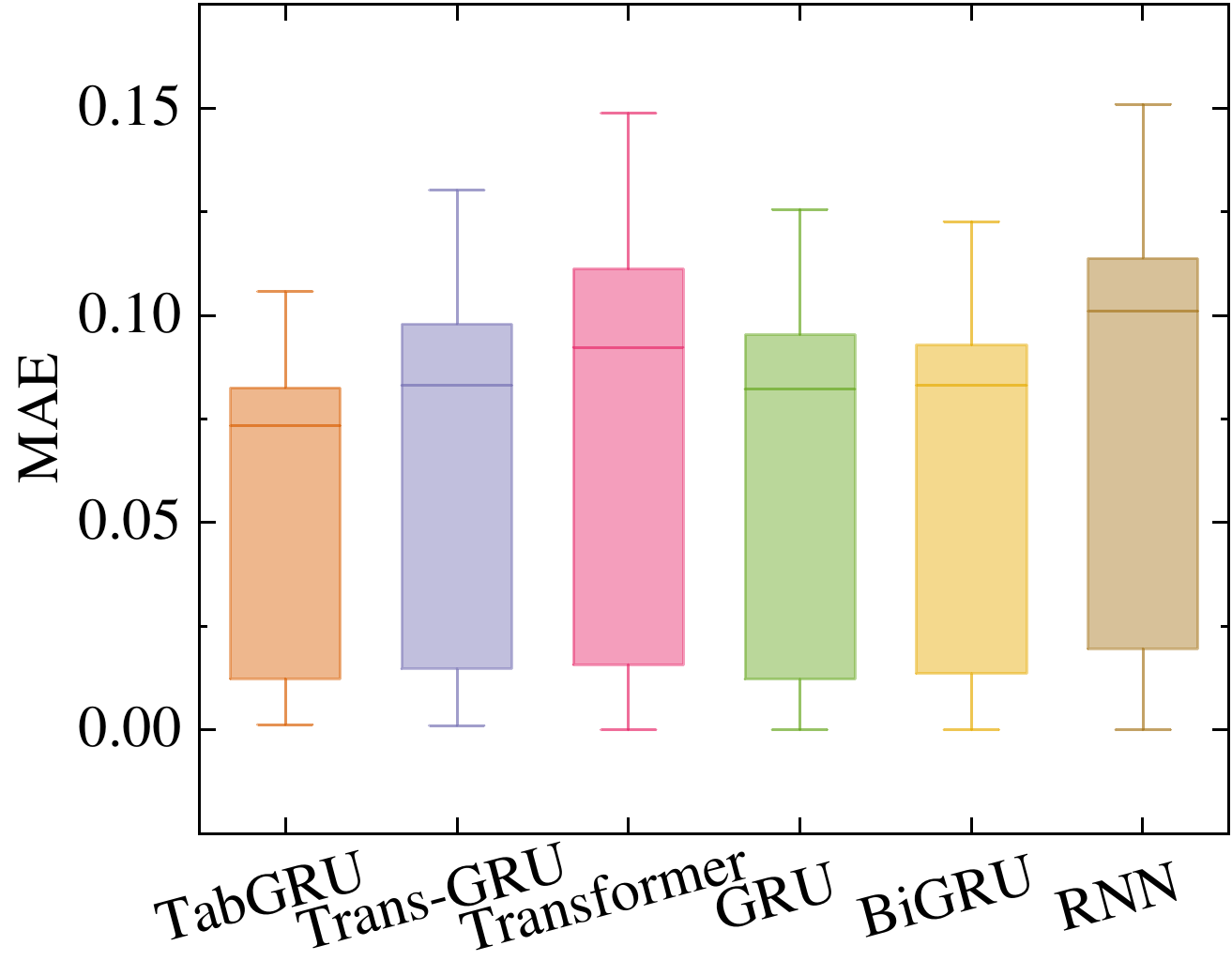}
    \label{fig:Torp_MAE}}
  
  \caption{\textcolor{blue}{Performance Evaluation Metrics on the Torp Site’s Test Set}}
  \label{Torpbox}
\end{figure}

\begin{figure}[htbp]
  \centering
  \subfloat[\textcolor{blue}{RMSE}]{
    \includegraphics[width=0.48\columnwidth]{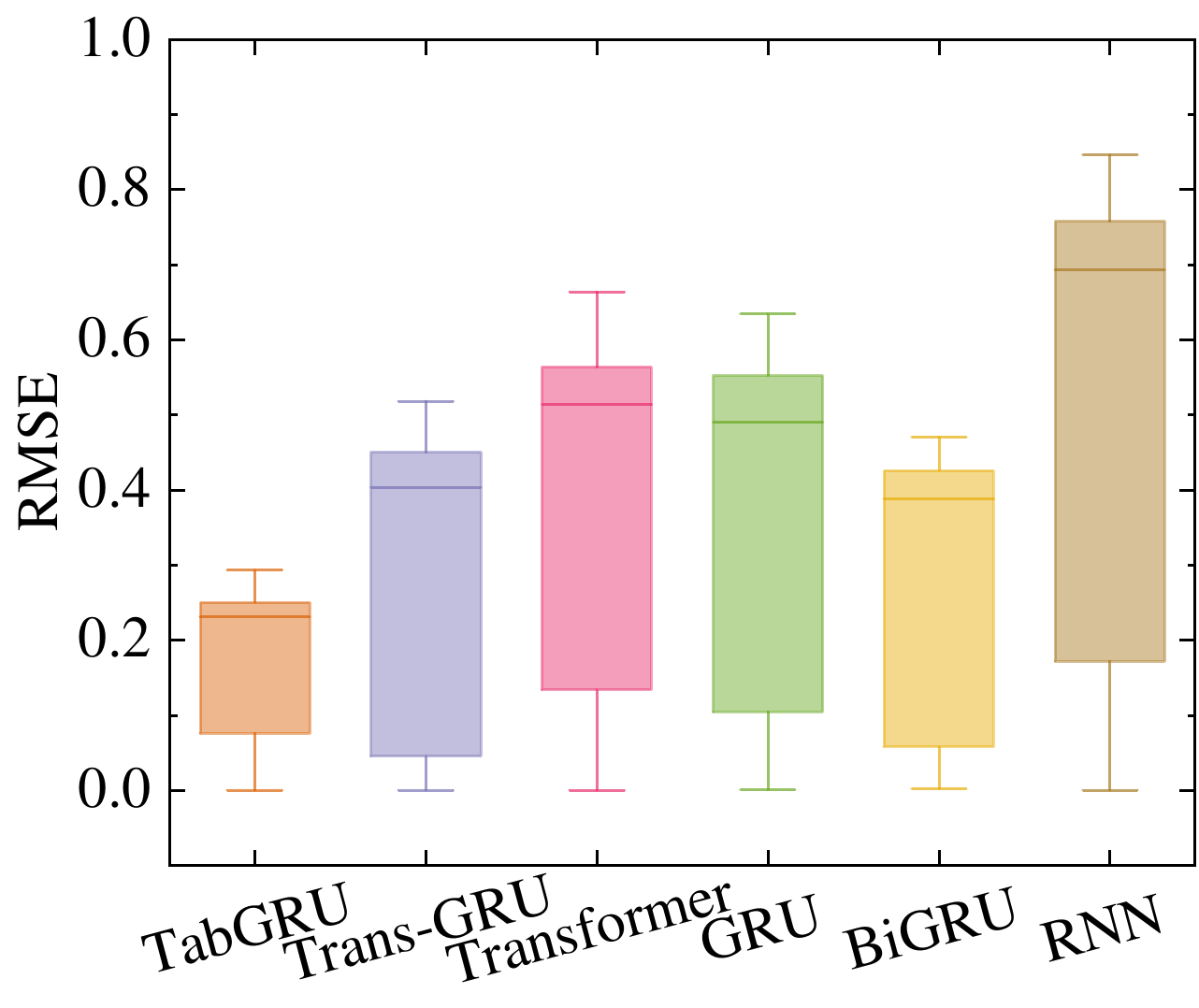}
    \label{fig:Barl_RMSE}}
  \hfill
  \subfloat[\textcolor{blue}{$R^2$}]{
    \includegraphics[width=0.48\columnwidth]{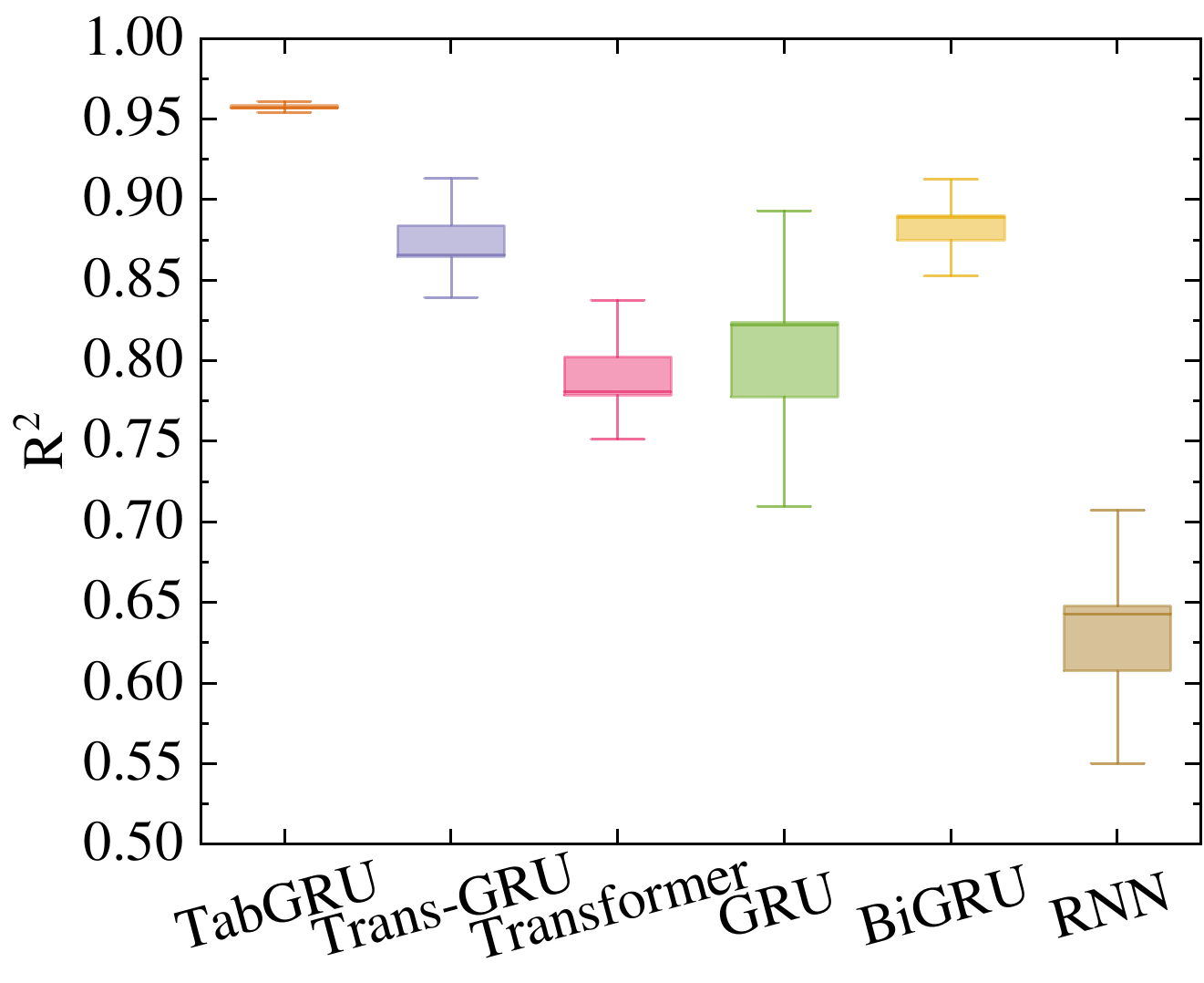}
    \label{fig:Barl_R2}}
  
  \vspace{0.2cm}
  
  \subfloat[\textcolor{blue}{PCC}]{
    \includegraphics[width=0.48\columnwidth]{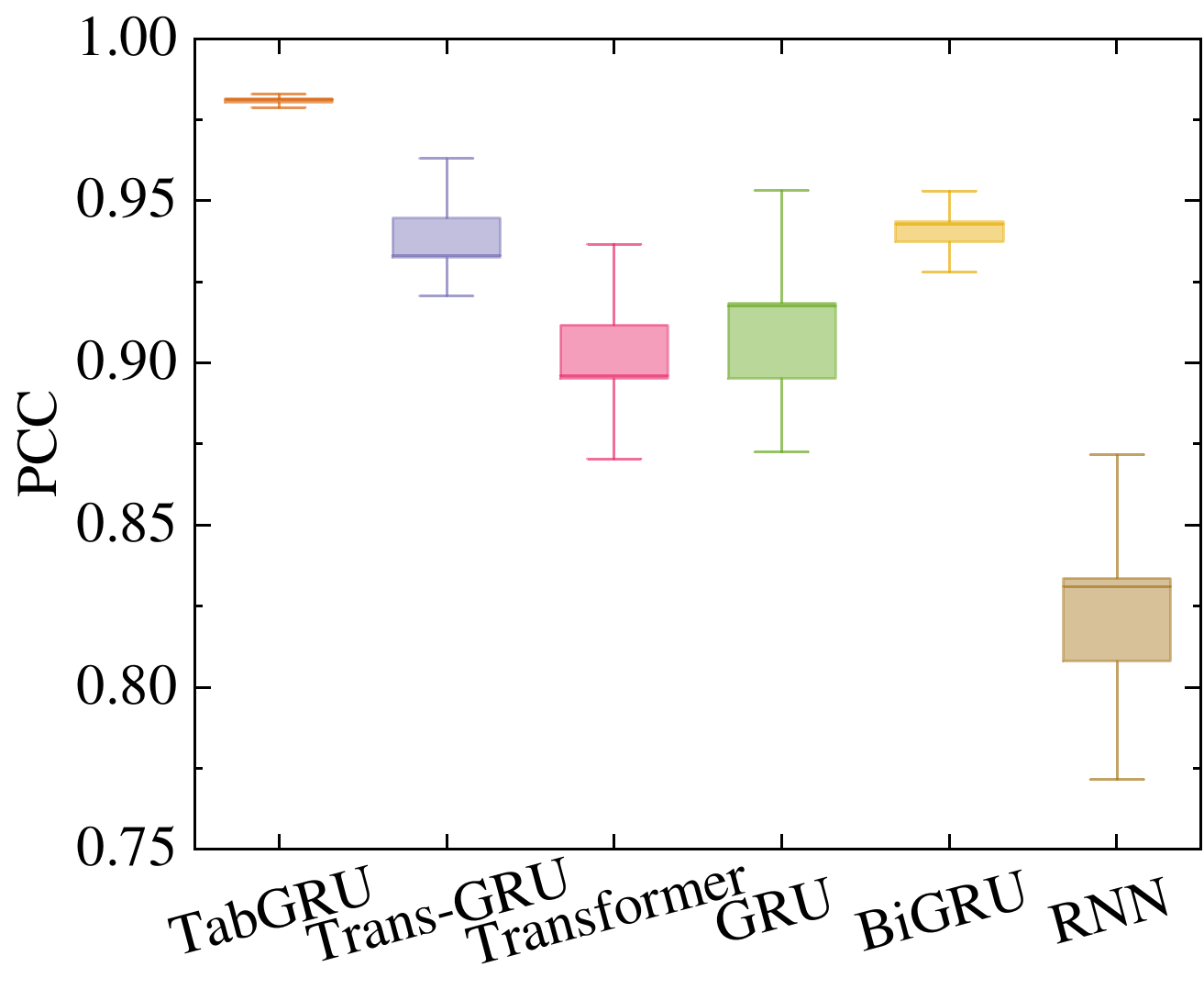}
    \label{fig:Barl_PCC}}
  \hfill
  \subfloat[\textcolor{blue}{MAE}]{
    \includegraphics[width=0.48\columnwidth]{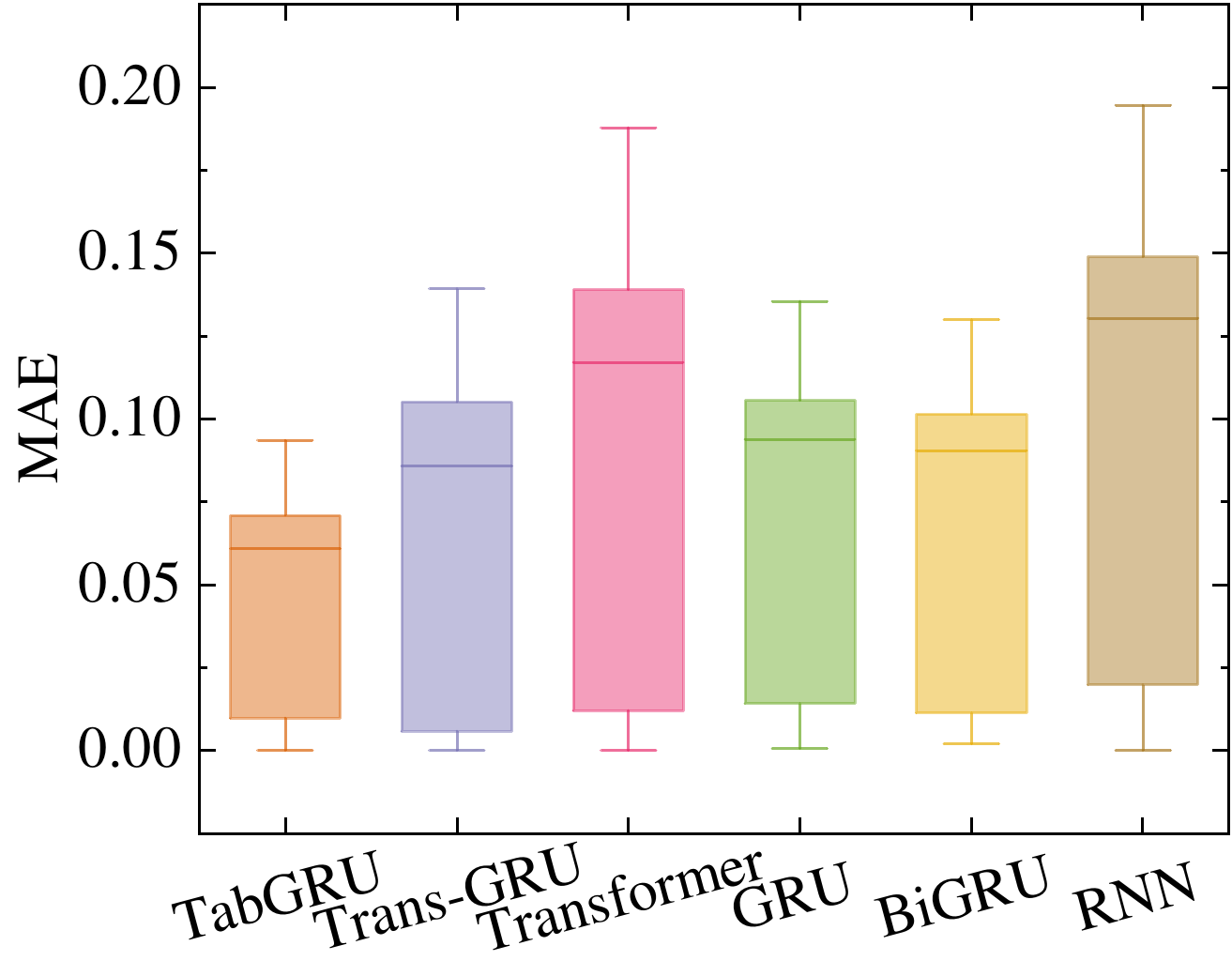}
    \label{fig:Barl_MAE}}
  
  \caption{\textcolor{blue}{Performance Evaluation Metrics on the Barl Site’s Test Set}}
  \label{Barlbox}
\end{figure}

\textcolor{blue}{In addition to the overall performance metrics, the model's performance consistency throughout the test period was examined. Figure \ref{Torpbox} and Figure \ref{Barlbox} visualize the distribution of daily performance metrics over the 10-day test period at both sites using box plots. Regarding error metrics (RMSE and MAE), TabGRU's median values were the lowest among all models on both the Torp and Barl datasets, and its interquartile range (IQR, i.e., the box height) was also comparatively the narrowest. This suggests that TabGRU's prediction errors were not only lower on average but also more stable across different days. In terms of goodness-of-fit (R$^2$ and PCC), the median values for TabGRU were stable around 0.90 and 0.96 at the Torp site, and around 0.96 and 0.98 at the Barl site. These values were higher than all baseline models while maintaining similarly compact distributions, confirming a strong predictive capability maintained consistently throughout the test period.}

\subsection{Performance during a light rainfall event}
\textcolor{blue}{To examine model performance under different rainfall scenarios, we selected two representative days from the test set for in-depth studies: a light rainfall day (August 26) and a moderate-intensity rainfall day (August 27).}

\textcolor{blue}{First, we examined the model's performance on a typical light rainfall day (August 26). This day was characterized by low-intensity but high-variability precipitation: the mean rainfall rate measured by the Torp gauge was only 0.11 mm/h, while the standard deviation reached 0.39 mm/h. These subtle and unstable rainfall patterns pose a considerable challenge to the model's sensitivity and precision. Table \ref{tab:Torp26} provides a quantitative summary of all models' performance metrics for this day. Compared with the strongest baseline (Trans-GRU), TabGRU reduced the RMSE by 11.11\% and the MAE by 25.00\%, while improving the R$^2$ by 21.95\% and the PCC by 7.94\%.}
\begin{table}[htbp]
  \centering
  \caption{\textcolor{blue}{Statistical Analysis of Rainfall Estimation Results at the Torp Site on August 26th}}
  \begin{tabular}{c|c|c|c|c}
    \hline
    \parbox[c]{1.5cm}{\centering \textcolor{blue}{Model}} &
    \parbox[c]{1cm}{\centering \textcolor{blue}{RMSE}} &
    \parbox[c]{1cm}{\centering \textcolor{blue}{$R^2$}} &
    \parbox[c]{1cm}{\centering \textcolor{blue}{PCC}} &
    \parbox[c]{1cm}{\centering \textcolor{blue}{MAE}} \\
    \hline\hline

    \textcolor{blue}{RNN} & \textcolor{blue}{0.14} & \textcolor{blue}{0.19} & \textcolor{blue}{0.63} & \textcolor{blue}{0.04} \\
    \hline

    \textcolor{blue}{Transformer} & \textcolor{blue}{0.10} & \textcolor{blue}{0.36} & \textcolor{blue}{0.79} & \textcolor{blue}{0.04} \\
    \hline

    \textcolor{blue}{GRU} & \textcolor{blue}{0.11} & \textcolor{blue}{0.35} & \textcolor{blue}{0.64} & \textcolor{blue}{0.04} \\
    \hline

    \textcolor{blue}{BiGRU} & \textcolor{blue}{0.12} & \textcolor{blue}{0.22} & \textcolor{blue}{0.65} & \textcolor{blue}{0.04} \\
    \hline

    \textcolor{blue}{Trans-GRU} & \textcolor{blue}{0.09} & \textcolor{blue}{0.41} & \textcolor{blue}{0.63} & \textcolor{blue}{0.04} \\
    \hline

    \textcolor{blue}{TabGRU} &
    \textcolor{blue}{\textbf{0.08}} &
    \textcolor{blue}{\textbf{0.50}} &
    \textcolor{blue}{\textbf{0.68}} &
    \textcolor{blue}{\textbf{0.03}} \\
    \hline
  \end{tabular}
  \label{tab:Torp26}
\end{table}

\begin{figure}[!htbp] 
  \centering
  \includegraphics[width=0.8\linewidth,height=0.4\textheight,keepaspectratio]{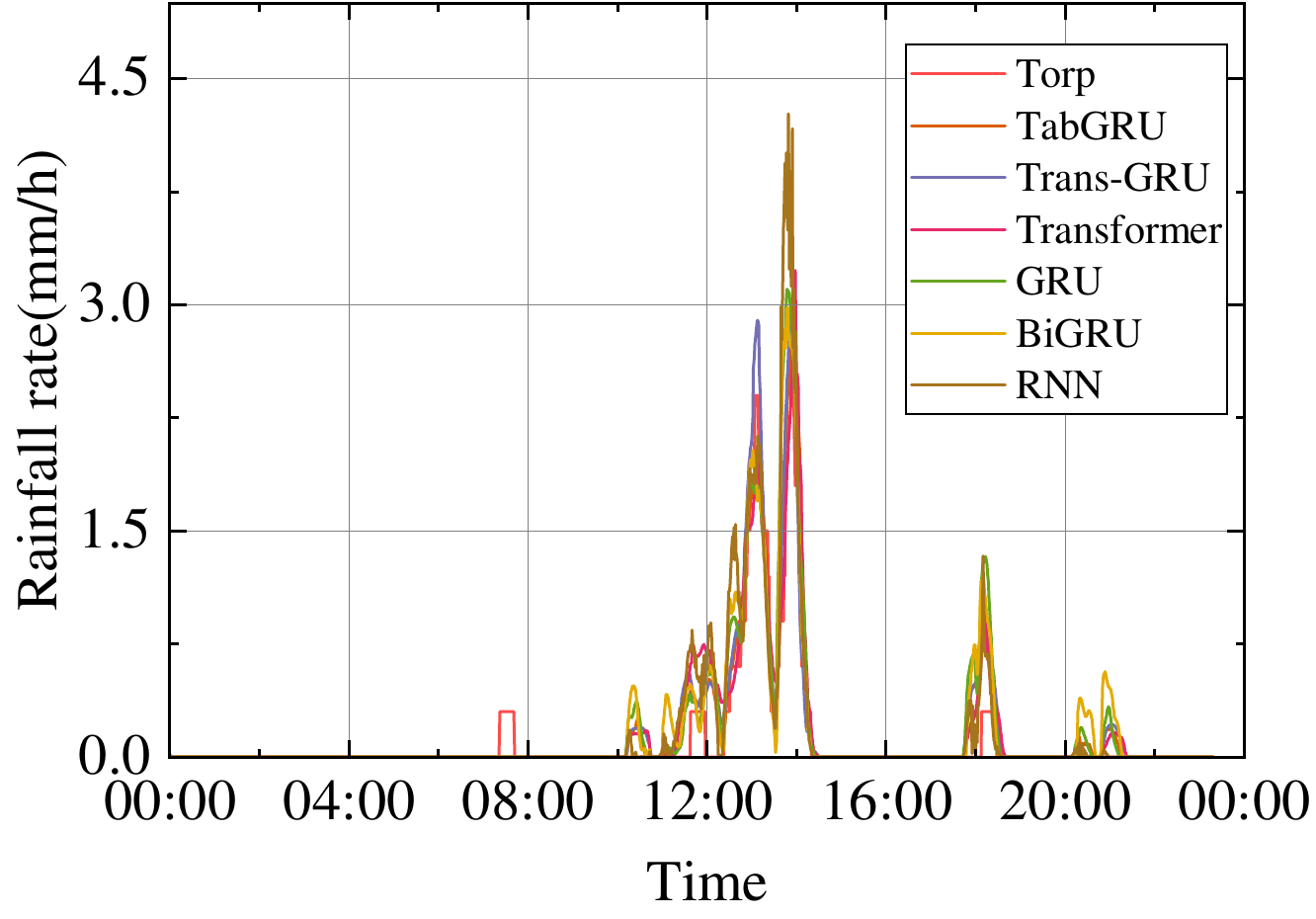} 
  \caption{\textcolor{blue}{Comparison Chart of Measured Rainfall Rates by Trop and Estimated Rainfall Rates by Various Models on August 26.}} 
  \label{fig:Torp26}
\end{figure}

\textcolor{blue}{The time-series comparison in Figure \ref{fig:Torp26} shows the main rainfall event occurred between 12:00 and 14:00, with a peak intensity of 2.70 mm/h. While most baseline models captured the general trend, they exhibited an overestimation bias. The RNN model, in particular, showed a pronounced overestimation of the peak. In contrast, the TabGRU prediction aligned closely with the ground-truth (Torp) peak (approx. 0.1 mm/h difference) and mitigated the overestimation bias observed in the other models during this event.}
\ref{fig:Torp26}.

\textcolor{blue}{On August 26th, the Barl rain gauge observed an average rainfall intensity of only 0.11 mm/h, while the standard deviation reached 0.41 mm/h, presenting a low-intensity, high-variability rainfall pattern. Table \ref{tab:Barl26} quantitatively summarizes the performance metrics of all models at the Barl site for this day. The TabGRU model's coefficient of determination (R$^2$ ) was the highest (0.65). In terms of error metrics, TabGRU (0.11 mm/h) jointly achieved the lowest RMSE with BiGRU (0.11 mm/h) and shared the lowest MAE (0.04 mm/h) with GRU and BiGRU. A noteworthy comparison is that the basic RNN model obtained the highest PCC (0.89), despite its relatively high RMSE (0.14 mm/h).}
\begin{table}[htbp]
  \centering
  \caption{\textcolor{blue}{Statistical Analysis of Rainfall Estimation Results at the Barl Site on August 26th}}
  \begin{tabular}{c|c|c|c|c}
    \hline
    \parbox[c]{1.5cm}{\centering \textcolor{blue}{Model}} & 
    \parbox[c]{1cm}{\centering \textcolor{blue}{RMSE}} & 
    \parbox[c]{1cm}{\centering \textcolor{blue}{$R^2$}} & 
    \parbox[c]{1cm}{\centering \textcolor{blue}{PCC}} & 
    \parbox[c]{1cm}{\centering \textcolor{blue}{MAE}} \\
    \hline\hline
    \textcolor{blue}{RNN} & \textcolor{blue}{0.14} & \textcolor{blue}{0.54} & \textcolor{blue}{\textbf{0.89}} & \textcolor{blue}{0.05} \\
    \hline
    \textcolor{blue}{Transformer} & \textcolor{blue}{0.16} & \textcolor{blue}{0.36} & \textcolor{blue}{0.68} & \textcolor{blue}{0.06} \\
    \hline
    \textcolor{blue}{GRU} & \textcolor{blue}{0.12} & \textcolor{blue}{0.64} & \textcolor{blue}{0.81} & \textcolor{blue}{\textbf{0.04}} \\
    \hline
    \textcolor{blue}{BiGRU} & \textcolor{blue}{\textbf{0.11}} & \textcolor{blue}{0.60} & \textcolor{blue}{0.80} & \textcolor{blue}{\textbf{0.04}} \\
    \hline
    \textcolor{blue}{Trans-GRU} & \textcolor{blue}{0.18} & \textcolor{blue}{0.22} & \textcolor{blue}{0.52} & \textcolor{blue}{0.06} \\
    \hline
    \textcolor{blue}{TabGRU} & \textcolor{blue}{\textbf{0.11}} & \textcolor{blue}{\textbf{0.65}} & \textcolor{blue}{0.80} & \textcolor{blue}{\textbf{0.04}} \\
    \hline
  \end{tabular}
  \label{tab:Barl26}
\end{table}

\begin{figure}[!htbp] 
  \centering
  \includegraphics[width=0.8\linewidth,height=0.4\textheight,keepaspectratio]{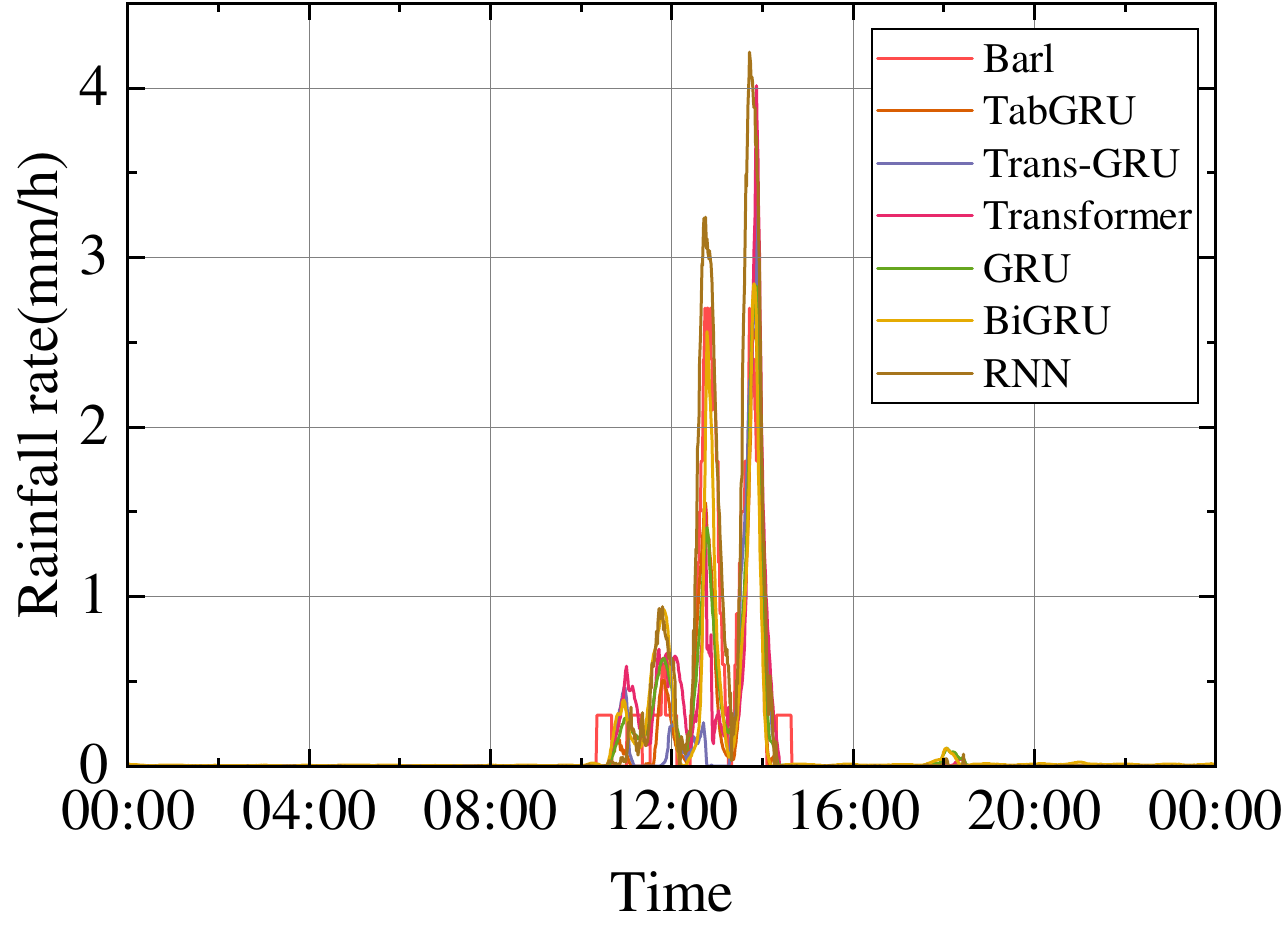} 
  \caption{\textcolor{blue}{Comparison Chart of Measured Rainfall Rates by Barl and Estimated Rainfall Rates by Various Models on August 26.}} 
  \label{fig:Barl26}
\end{figure}

\textcolor{blue}{The time-series comparison in Figure \ref{fig:Barl26} provides a direct visual corroboration for the quantitative metrics at the Barl site. It can be seen that the rainfall was mainly concentrated between 11:00 and 15:00, and reached a peak of approximately 2.7 mm/h twice. Among the models, the fluctuation rhythm of the basic RNN model's prediction curve was highly consistent with the ground truth, but it exhibited a significant overestimation problem, with its predicted second peak, approximately 4.2 mm/h, being much higher than the ground truth. In contrast, the TabGRU model's prediction curve showed the minimal deviation from the ground truth, especially at the second peak, where its estimated value (approx. 2.5 mm/h) was very close to the true value. Furthermore, the deep learning models exhibit a systematic bias: underestimation at the beginning of a rainfall event and overestimation at the end. This phenomenon is due to the hysteresis effect of WAA. This is because, at the onset of rain, path attenuation has already occurred, but the antenna radome itself is still dry. When the model sees this lower-than-expected total attenuation, it may underestimate it. Conversely, when the rain stops, the path attenuation drops to zero, but the residual water film on the antenna surface may cause the model to overestimate it as light rain.}

\subsection{Performance during a moderate intensity rainfall event}
\textcolor{blue}{Next, we evaluated TabGRU's performance during a moderate-intensity rainfall event on August 27. The Torp rain gauge observed an average rainfall intensity of 0.97 mm/h, while the standard deviation reached 2.09 mm/h. Table \ref{tab:Torp27} provides a quantitative summary of all models' performance metrics for this event. It can be observed that during this event, the TabGRU model achieved the optimal values across all four evaluation metrics. Compared with the top-performing baseline model (BiGRU), its RMSE and MAE were reduced by 16.46\% and 13.33\%, respectively, while the R$^2$ and PCC were improved by 1.16\% and 2.08\%, respectively.}
\begin{table}[htbp]
  \centering
  \caption{\textcolor{blue}{Statistical Analysis of Rainfall Estimation Results at the Torp Site on August 27th}}
  \begin{tabular}{c|c|c|c|c}
   \hline
   \parbox[c]{1.5cm}{\centering \textcolor{blue}{Model}} & 
   \parbox[c]{1cm}{\centering \textcolor{blue}{RMSE}} & 
   \parbox[c]{1cm}{\centering \textcolor{blue}{$R^2$}} & 
   \parbox[c]{1cm}{\centering \textcolor{blue}{PCC}} & 
   \parbox[c]{1cm}{\centering \textcolor{blue}{MAE}} \\
    \hline\hline
    \textcolor{blue}{RNN} & \textcolor{blue}{1.32} & \textcolor{blue}{0.67} & \textcolor{blue}{0.89} & \textcolor{blue}{0.44} \\
    \hline
    \textcolor{blue}{Transformer} & \textcolor{blue}{1.37} & \textcolor{blue}{0.64} & \textcolor{blue}{0.83} & \textcolor{blue}{0.49} \\
    \hline
    \textcolor{blue}{GRU} & \textcolor{blue}{1.09} & \textcolor{blue}{0.76} & \textcolor{blue}{0.90} & \textcolor{blue}{0.38} \\
    \hline
    \textcolor{blue}{BiGRU} & \textcolor{blue}{0.79} & \textcolor{blue}{0.86} & \textcolor{blue}{0.96} & \textcolor{blue}{0.30} \\
    \hline
    \textcolor{blue}{Trans-GRU} & \textcolor{blue}{1.03} & \textcolor{blue}{0.76} & \textcolor{blue}{0.90} & \textcolor{blue}{0.38} \\
    \hline
    \textcolor{blue}{TabGRU} & \textcolor{blue}{\textbf{0.66}} & \textcolor{blue}{\textbf{0.87}} & \textcolor{blue}{\textbf{0.98}} & \textcolor{blue}{\textbf{0.26}} \\
    \hline
  \end{tabular}
  \label{tab:Torp27}
\end{table}

\begin{figure}[!htbp] 
  \centering
  \includegraphics[width=0.8\linewidth,height=0.4\textheight,keepaspectratio]{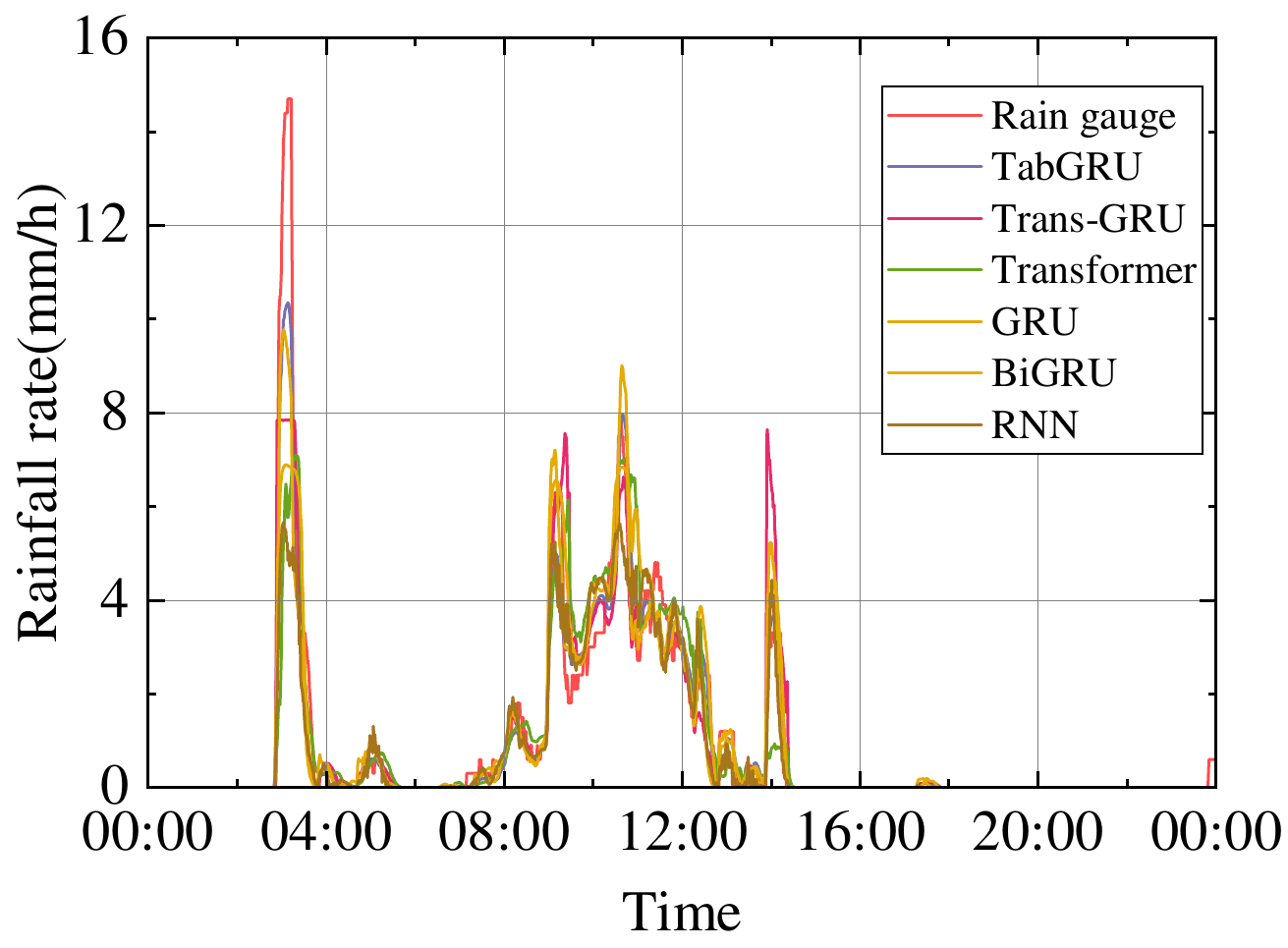} 
  \caption{\textcolor{blue}{Comparison Chart of Measured Rainfall Rates by Trop and Estimated Rainfall Rates by Various Models on August 27.}} 
  \label{fig:Torp27}
\end{figure}

\textcolor{blue}{The time-series comparison in Figure \ref{fig:Torp27} provides a direct visual corroboration for this result. The event contained multiple peaks, notably an extreme peak of approximately 15 mm/h around 03:00. All models exhibited underestimation at this point. Among them, TabGRU's predicted peak was approximately 10.20 mm/h, while BiGRU's peak was approximately 9.7 mm/h. In the subsequent rainfall events from 08:30 to 12:00, TabGRU's prediction curve also closely matched the morphology of the ground truth. This visual performance aligns consistently with TabGRU's results in the quantitative table, where it achieved the lowest RMSE (0.66 mm/h) and the highest PCC (0.98).}

\textcolor{blue}{The Barl rain gauge observed an average rainfall intensity of 1.01 mm/h, while the standard deviation reached 2.04 mm/h. Table \ref{tab:Barl27} provides a quantitative summary of all models' performance metrics for this event. It can be observed that during this event, the TabGRU model achieved the optimal values across all four evaluation metrics. Compared with the top-performing baseline model (GRU), its RMSE and MAE were reduced by 29.03\% and 20.69\%, respectively, while the R$^2$ and PCC were improved by 8.14\% and 3.16\%, respectively.}
\begin{table}[htbp]
  \centering
  \caption{\textcolor{blue}{Statistical Analysis of Rainfall Estimation Results at the Barl Site on August 27th}}
  \begin{tabular}{c|c|c|c|c}
   \hline
   \parbox[c]{1.5cm}{\centering \textcolor{blue}{Model}} & 
   \parbox[c]{1cm}{\centering \textcolor{blue}{RMSE}} & 
   \parbox[c]{1cm}{\centering \textcolor{blue}{$R^2$}} & 
   \parbox[c]{1cm}{\centering \textcolor{blue}{PCC}} & 
   \parbox[c]{1cm}{\centering \textcolor{blue}{MAE}} \\
    \hline\hline
    \textcolor{blue}{RNN} & \textcolor{blue}{0.96} & \textcolor{blue}{0.70} & \textcolor{blue}{0.88} & \textcolor{blue}{0.42} \\
    \hline
    \textcolor{blue}{Transformer} & \textcolor{blue}{1.08} & \textcolor{blue}{0.55} & \textcolor{blue}{0.84} & \textcolor{blue}{0.53} \\
    \hline
    \textcolor{blue}{GRU} & \textcolor{blue}{0.62} & \textcolor{blue}{0.86} & \textcolor{blue}{0.95} & \textcolor{blue}{0.29} \\
    \hline
    \textcolor{blue}{BiGRU} & \textcolor{blue}{0.64} & \textcolor{blue}{0.83} & \textcolor{blue}{0.95} & \textcolor{blue}{0.32} \\
    \hline
    \textcolor{blue}{Trans-GRU} & \textcolor{blue}{0.93} & \textcolor{blue}{0.39} & \textcolor{blue}{0.92} & \textcolor{blue}{0.44} \\
    \hline
    \textcolor{blue}{TabGRU} & \textcolor{blue}{\textbf{0.44}} & \textcolor{blue}{\textbf{0.93}} & \textcolor{blue}{\textbf{0.98}} & \textcolor{blue}{\textbf{0.23}} \\
    \hline
  \end{tabular}
  \label{tab:Barl27}
\end{table}

\begin{figure}[!htbp] 
  \centering
  \includegraphics[width=0.8\linewidth,height=0.4\textheight,keepaspectratio]{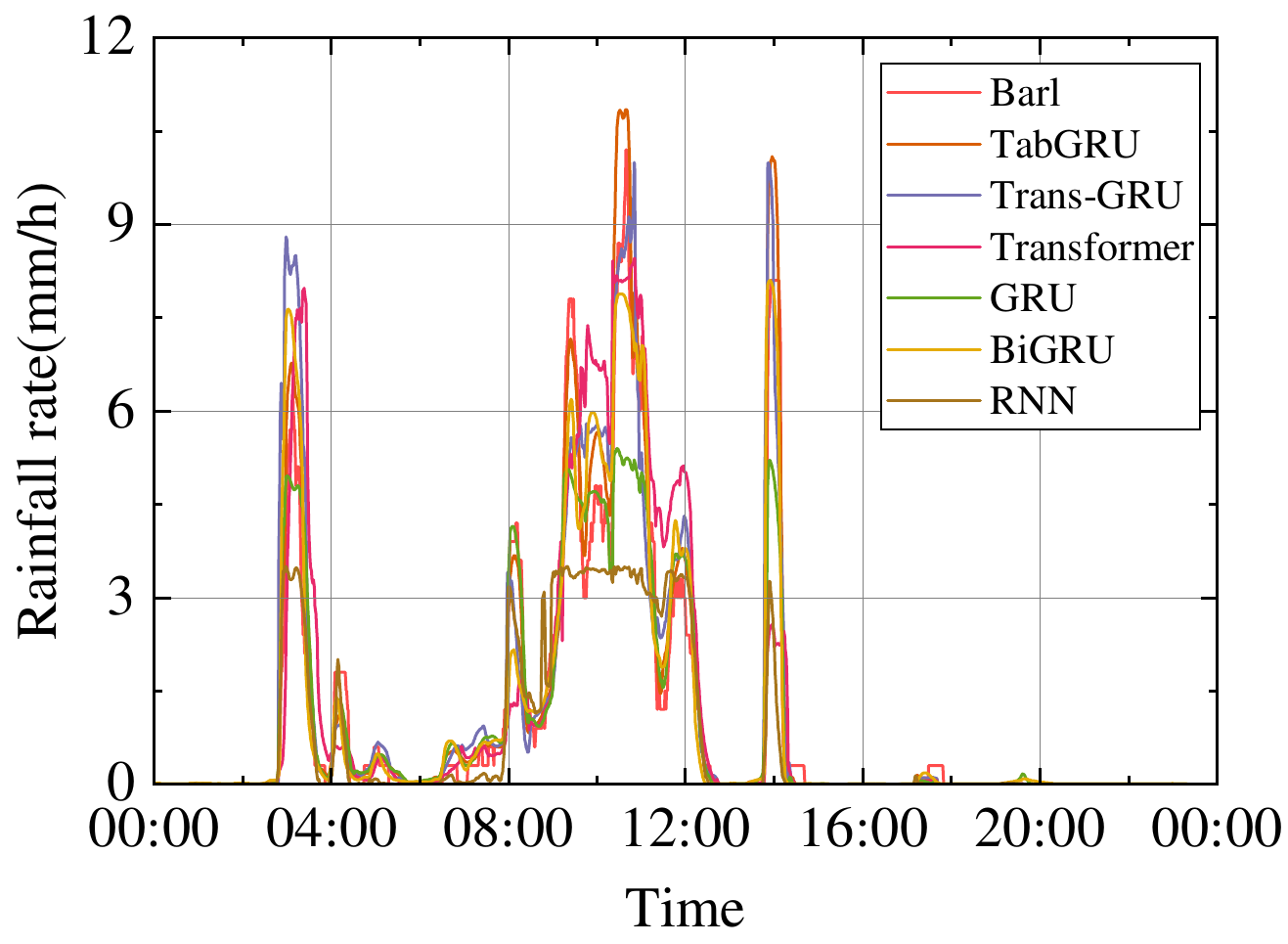} 
  \caption{\textcolor{blue}{Comparison Chart of Measured Rainfall Rates by Barl and Estimated Rainfall Rates by Various Models on August 27.} }
  \label{fig:Barl27}
\end{figure}

\textcolor{blue}{The time-series comparison in Figure \ref{fig:Barl27} provides a direct visual corroboration for this result. It can be seen from the plot that the rainfall event between 08:00 and 12:00 contained multiple peaks, with the highest peak appearing around 10:30 at approximately 10.80 mm/h. At this point, TabGRU's predicted value was approximately 10.20 mm/h, while GRU's predicted value was only about 5.36 mm/h. During other time periods, TabGRU's prediction curve was highly consistent with the ground truth, whereas other models (such as RNN) exhibited significant underestimation problems. This lower peak error and morphological consistency demonstrated by TabGRU aligns with the results in the quantitative table, where it achieved the highest PCC (0.98) and the lowest RMSE (0.44 mm/h).}

\subsection{Comparison with the model-driven approach}
The preceding comparative analysis establishes the superiority of TabGRU among the tested deep learning architectures. To further and more comprehensively evaluate the advantage of the data-driven approach, this section presents a direct comparison between TabGRU and the classical PL model. This model-driven approach is adapted from the research framework of Zheng et al. (2021)\cite{s21051670}, which similarly utilizes the increase in the standard deviation of the microwave signal during rainfall to classify wet and dry periods and applies the PL relationship recommended by the ITU-R to convert rain-induced attenuation into rainfall rate. As the dataset used in this paper lacks DSD data, we implemented a practical simplification for the WAA correction: a fixed attenuation offset, estimated solely from the microwave link's own frequency information, is used as an approximate correction value for WAA. In the following, we will conduct a comprehensive performance examination of these two methods across multiple evaluation metrics.

\textcolor{blue}{Table \ref{tab:TorpPL} presents the comparative statistical results for TabGRU and the PL model on the entire test set at the Torp site. The quantitative data clearly indicate that, compared to the PL model, TabGRU's RMSE and MAE were reduced by 27.66\% and 27.27\%, respectively, while the R$^2$ and PCC were improved by 9.64\% and 4.35\%, respectively. }
\begin{table}[htbp]
  \centering
  \caption{\textcolor{blue}{Statistical analysis of rainfall estimates from the PL model and our model at the Torp site}}
  {\color{blue} 
  \begin{tabular}{c|c|c|c|c}
   \hline
   \parbox[c]{1.5cm}{\centering Model} & 
   \parbox[c]{1cm}{\centering RMSE} & 
   \parbox[c]{1cm}{\centering $R^2$} & 
   \parbox[c]{1cm}{\centering PCC} & 
   \parbox[c]{1cm}{\centering MAE} \\
    \hline\hline
    PL model & 0.47 & 0.83 & 0.92 & 0.11 \\
    \hline
    TabGRU & \textbf{0.34} & \textbf{0.91} & \textbf{0.96} & \textbf{0.08} \\
    \hline
  \end{tabular}
  }
  \label{tab:TorpPL}
\end{table}

\textcolor{blue}{Table \ref{tab:BarlPL} presents the comparative statistical results for TabGRU and the PL model on the entire test set at the Barl site. The quantitative data clearly indicate that, compared to the PL model, TabGRU's RMSE and MAE were reduced by 30.56\% and 45.45\%, respectively, while the R$^2$ and PCC were improved by 9.09\% and 2.08\%, respectively.}
\begin{table}[htbp]
  \centering
  \caption{\textcolor{blue}{Statistical analysis of rainfall estimates from the PL model and our model at the Barl site}}
  {\color{blue} 
  \begin{tabular}{c|c|c|c|c}
   \hline
   \parbox[c]{1.5cm}{\centering Model} & 
   \parbox[c]{1cm}{\centering RMSE} & 
   \parbox[c]{1cm}{\centering $R^2$} & 
   \parbox[c]{1cm}{\centering PCC} & 
   \parbox[c]{1cm}{\centering MAE} \\
    \hline\hline
    PL model & 0.36 & 0.88 & 0.96 & 0.11 \\
    \hline
    TabGRU & \textbf{0.25} & \textbf{0.96} & \textbf{0.98} & \textbf{0.06} \\
    \hline
  \end{tabular}
  }
  \label{tab:BarlPL}
\end{table}

\begin{figure}[!htbp] 
  \centering
  \includegraphics[width=0.9\linewidth,height=0.4\textheight,keepaspectratio]{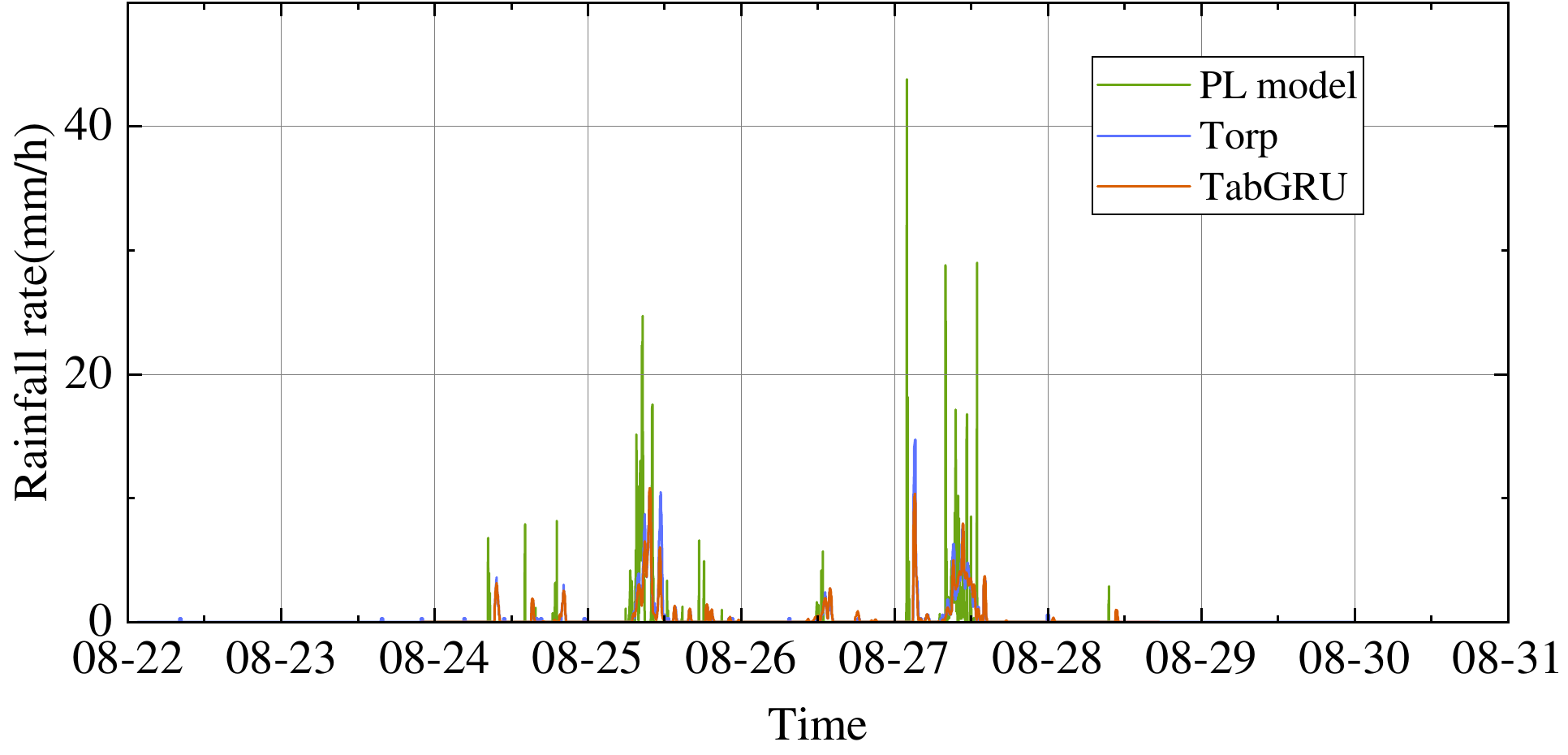} 
  \caption{\textcolor{blue}{Time sequence comparison of rainfall rates from the proposed model, the PL model, and the rain gauge at the Torp site during the testing period (August 22 to 31).}} 
  \label{fig:TorpPL}
\end{figure}

\begin{figure}[!htbp] 
  \centering
  \includegraphics[width=0.9\linewidth,height=0.4\textheight,keepaspectratio]{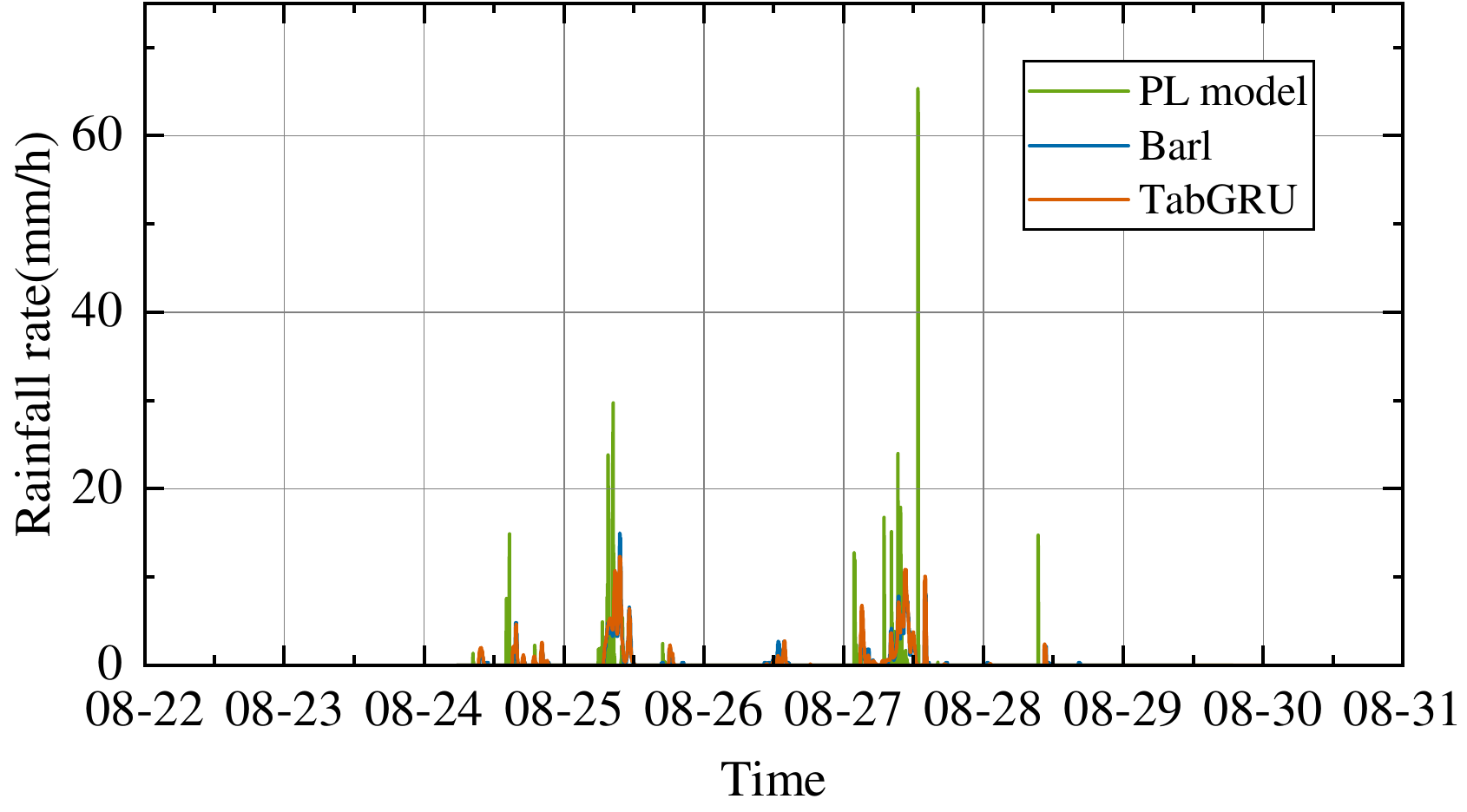} 
  \caption{\textcolor{blue}{Time sequence comparison of rainfall rates from the proposed model, the PL model, and the rain gauge at the Barl site during the testing period (August 22 to 31).}} 
  \label{fig:BarlPL}
\end{figure}

\textcolor{blue}{Figure \ref{fig:TorpPL} and Figure \ref{fig:BarlPL} compare the temporal performance of TabGRU and the PL model at the Torp and Barl site, respectively, during the testing period. As can be seen from the two plots, the primary difference between the TabGRU and PL models is manifested during high-intensity rainfall events. At both sites, the PL model exhibits a significant overestimation problem. In contrast, the TabGRU model's deviation from the ground truth during these peak events is substantially smaller than that of the PL model. Although it also shows some underestimation at the Torp site, it almost perfectly reproduces the true value at the Barl site.}

In summary, whether considering the quantitative statistics of overall performance or predictive efficacy in specific rainfall events, the proposed deep learning model demonstrates comprehensive superiority. It not only achieves significantly higher predictive accuracy than the traditional physical model and other deep learning baselines but also exhibits a high degree of stability and reliability across diverse rainfall scenarios. These results provide compelling evidence that the data-driven approach holds a clear advantage for the complex task of CML-based rainfall estimation tasks.

\section{Conclusion}
To improve the accuracy of high-resolution spatiotemporal rainfall estimation in urban areas, this paper proposes and validates a deep learning method for rainfall estimation using CMLs—TabGRU. The model enhances the standard Transformer by introducing a BiGRU layer, optimizing its capability to capture localized, short-term rainfall fluctuations and enabling it to model both long-term trends and short-term variations simultaneously. Furthermore, TabGRU employs a customized learnable positional encoding and an attention pooling mechanism to enhance dynamic feature extraction and improve generalization for complex rainfall sequences.

\textcolor{blue}{The effectiveness of TabGRU was validated on a public benchmark dataset from Gothenburg, Sweden (June-September 2015), using data from the Torp and Barl rain gauges and 12 sub-links. The experimental results show a consistent performance advantage: TabGRU outperformed the other deep learning models used in this study at both the Torp and Barl sites. While RNN models like BiGRU and GRU can effectively capture local temporal features, TabGRU's hybrid architecture design allows it to model long-range dependencies while simultaneously handling fine-grained local fluctuations, thus optimizing prediction accuracy. Furthermore, this paper compared TabGRU with the PL model; TabGRU performed well at both sites, whereas the PL model exhibited significant overestimation, especially near rainfall peaks. These results indicate that, under the test conditions of this paper, TabGRU is a robust and effective solution for the task of CML-based rainfall estimation.}

\textcolor{blue}{While the results are promising, this study also identifies key directions for future optimization. First, building on the two-site validation in this study, future work will involve testing on new datasets from different geographical and climatological regions to enhance model generalization. Second, to improve the estimation capability for extremely high-intensity rainfall events, future work will focus on incorporating more diverse datasets containing such events. Third, this study found that the precise quantification of micro-precipitation (light/drizzle) events remains a challenge—as shown in specific cases (e.g., Barl site, August 26, 2015), simpler baseline models like BiGRU occasionally achieved lower errors in these low-signal scenarios. Therefore, future work will focus on improving model sensitivity in the low-intensity rainfall spectrum. This includes integrating a more robust wet/dry classification module and exploring techniques such as focal loss or specialized sub-models to better handle the class imbalance and noise interference present in drizzle event data. Finally, future models should also explore state-aware architectures to address the systematic bias of underestimation at the onset of rainfall and overestimation at the end. This will allow the model to effectively distinguish between the antenna's wetting/drying phases and the true path attenuation. These efforts will collectively enhance the model's robustness and fully tap its potential for urban meteorological monitoring.}

\section{Acknowledgments}
The authors acknowledge support from the National Key R \& D Program of China (2023YFC3010700), the Basic Research Program of Jiangsu (BK20231148), and the National Natural Science Foundation of China (\textcolor{blue}{42575153},42027803, 42575153).

\bibliographystyle{ieeetr}
\bibliography{Ref.bib}

\end{document}